\documentclass{article}  

\usepackage[a4paper, margin=2.5cm]{geometry}
\usepackage{graphicx}%
\usepackage[numbers]{natbib}
\usepackage{multirow}%
\usepackage{amsmath,amssymb,amsfonts}%
\usepackage{amsthm}%
\usepackage{mathrsfs}%
\usepackage[title]{appendix}%
\usepackage{xcolor}%
\usepackage{textcomp}%
\usepackage{manyfoot}%
\usepackage{booktabs}%
\usepackage{algorithm}%
\usepackage{algpseudocode}
\usepackage{listings}%
\usepackage{array}%
\usepackage[colorlinks=true, linkcolor=blue, citecolor=blue, urlcolor=blue]{hyperref}
\usepackage{threeparttable}

\algrenewcommand\algorithmicdo{\textcolor{blue}{\textbf{do}}}
\algrenewcommand\algorithmicend{\textcolor{red}{\textbf{end}}}
\algrenewcommand\algorithmicfor{\textcolor{green}{\textbf{for}}}

\title{Addressing Hallucinations with RAG and NMISS in Italian Healthcare LLM Chatbots}

\author{
    \parbox{\textwidth}{\centering
        \textbf{Maria Paola Priola*} \\ \vspace{1em}
        \texttt{mariap.priola@unica.it} \\ \vspace{1em}
        {\small *Department of Economics and Business, University of Cagliari} \\ 
        {\small Via Sant'Ignazio da Laconi, 17, 09123 Cagliari, Italy}
    }
}

\date{}

\begin{document}

\maketitle 

    \begin{abstract}
        This work addresses hallucinations in Large Language Models (LLMs) by combining detection and mitigation strategies within a knowledge-driven evaluation framework. I employ a Retrieval-Augmented Generation (RAG) setup to ground model responses in structured external data, thereby improving factual reliability and helping models better follow instructions in zero-shot prompting settings. To enhance evaluation beyond surface-level overlap, I introduce the Negative Missing Information Scoring System (NMISS), a novel scoring system that complements traditional metrics such as BLEU, ROUGE, and METEOR by incorporating contextual relevance. NMISS helps distinguish between genuine hallucinations and responses that correctly infer missing information from trusted sources. Applied to a health-related question-answering task using Italian news articles, NMISS reveals important differences in model behavior. While Gemma2 and GPT-4 produce the most reference-aligned answers under standard metrics, Gemma2 also achieves the highest NMISS outperformance, suggesting that its semantically rich outputs are often under-recognized by traditional scores. In contrast, GPT-4 gains less from NMISS due to its strict adherence to prompt phrasing, leaving little room for contextual enhancement. Mid-tier models such as LLaMA2, LLaMA3, and Mistral show substantial NMISS gains when scored with BLEU and ROUGE-1, indicating their ability to provide contextually meaningful content even when traditional scores are lower. Overall, the RAG–NMISS framework improves hallucination detection, reveals context-aware model behavior often overlooked by traditional metrics, and supports more reliable evaluation in real-world structured knowledge tasks.
    \end{abstract}
    

    \section{Introduction}\label{ch2:sec:introduction}
    
        Large Language Models (LLMs) are AI systems typically built on the Transformer architecture, as introduced by \citet{vaswani2017attention}. Designed to understand and generate human language, LLMs are trained on extensive text datasets, allowing them to grasp the features of language. These models perform various tasks, from basic text classification to advanced text generation, and more recent iterations can even interpret images and generate code.
    
        LLMs have surged in popularity due to their intuitive, prompt-based interfaces, enabling effortless interaction. The past few years have seen the continuous development of notable LLMs like ChatGPT \citep{OpenAI2022, achiam2023gpt}, Gemini \citep{Google2023}, Falcon \citep{penedo2023refinedweb}, Llama \citep{touvron2023llama, touvron2023llama2, meta2024llama3}, and Gemma \citep{team2024gemma,team2024gemma2}. As highlighted by \citet{huang2023survey}, these models excel in language comprehension \citep{hendrycks2020measuring, huang2024c}, text summarization \citep{zhang2024benchmarking}, and reasoning \citep{wei2023simple, kojima2022large, qiao2022reasoning}. As a result, their applications span various industries, such as processing electronic medical records, matching clinical trials, and drug discovery in healthcare. In finance, LLMs assist with fraud detection and trading, while in business, they automate customer service via chatbots and virtual assistants \citep{ozdemir2023quick}.
    
        However, LLMs face a significant challenge: hallucinations \citep{guerreiro2023hallucinations, varshneystitch}. Hallucinations occur when models generate nonsensical or inaccurate content, failing to reflect the source material accurately \citep{ji2023survey}. This behavior is a serious threat since it affects the application of LLMs to real-world cases \citep{kaddour2023challenges}, as hallucinations can undermine the reliability of their outputs. \cite{huang2023survey} classifies hallucinations into two types: factuality hallucinations, where generated content diverges from verifiable facts, and faithfulness hallucinations, where the content strays from user instructions or input context. Factuality hallucinations manifest as either factual inconsistencies or fabrications. For example, a model might mistakenly claim that ``The Colosseum is located in Berlin, Germany'' instead of Rome or invent details about non-existent technologies. Faithfulness hallucinations occur when the output deviates from user instructions or context. These can involve instruction inconsistencies (e.g., responding in Spanish when asked to translate into Italian), context inconsistencies (introducing unrelated information), or logical errors (e.g., miscalculating ``two times four'' as three).
    
        The mitigation of hallucinations has become an urgent concern due to the widespread use of LLMs in sensitive areas such as healthcare. Addressing hallucinations is challenging due to LLMs' reliance on reinforcement learning from human feedback \citep{ouyang2022training}, their opaque black-box nature that hides knowledge boundaries \citep{sun2023moss, ren2023investigating}, and their training on large datasets that may contain outdated or incorrect information \citep{zhang2023siren}. Additionally, their versatility across many tasks complicates mitigation efforts, while their performance makes hallucinations, especially factual fabrications, harder to detect.

        Researchers have explored several strategies to mitigate hallucinations. One promising approach is Retrieval-Augmented Generation (RAG) \citep{lewis2020rag}, which incorporates external knowledge during generation to ground outputs in factual information. RAG improves both the accuracy and timeliness of model responses by leveraging authoritative sources, producing verifiable answers, and increasing user confidence \citep{kang2023ever}. Other techniques include prompt engineering, which optimizes outputs via instruction design \citep{tonmoy2024comprehensive, white2023prompt}; self-refinement, where feedback on prior outputs enhances subsequent responses \citep{madaan2023selfrefineiterativerefinementselffeedback}; and prompt tuning, which adapts instruction embeddings during fine-tuning to improve task-specific performance \citep{lester-etal-2021-power}.
        
        Detecting hallucinations is a prerequisite for effective mitigation. Traditional metrics such as ROUGE \citep{lin2004rouge}, BLEU \citep{papineni2002bleu}, and METEOR \citep{banerjee2005meteor} measure lexical overlap between generated and reference text but often fail to capture semantic equivalence \citep{celikyilmaz2020evaluation}. More recent embedding-based metrics, such as BERTScore \citep{zhang2019bertscore}, use contextual embeddings to assess semantic similarity. Another direction involves “opening the box”—analyzing token probabilities or logit outputs to detect hallucinations \citep{azaria2023internal, varshneystitch}—though this approach requires access to internal model states, which is not feasible for proprietary systems.

        This work proposes a general, domain-independent framework for hallucination detection and mitigation in LLMs. For mitigation, a RAG architecture integrates external knowledge into the generation process, ensuring that outputs remain grounded in factual information. The methodology employs a multilingual Sentence Transformer model \citep{reimers-2019-sentence-bert, reimers-2020-multilingual-sentence-bert} to construct and query a vector database for contextual retrieval. A range of LLMs, specifically Gemma \citep{team2024gemma}, Gemma-2 \citep{team2024gemma2}, Mistral \citep{jiang2023mistral}, GPT-4 \citep{achiam2023gpt}, Llama-2 \citep{touvron2023llama2}, and Llama-3 \citep{meta2024llama3} are compared using a zero-shot prompting strategy that instructs models to generate answers based on retrieved context. To evaluate interpretability and task complexity, 100 questions are categorized into three difficulty levels: low, medium, and high.

        For detection, this work introduces the \textbf{Negative Missing Information Scoring System (NMISS)}, a novel evaluation layer that extends traditional metrics by incorporating contextual relevance. Unlike prior metrics, which assess only lexical or semantic overlap with reference answers, NMISS leverages context from external sources to distinguish genuine hallucinations from contextually valid elaborations. Specifically, it examines lexical items present in model outputs but absent from human references, evaluating whether these additions are substantiated by the provided context. Through this mechanism, NMISS refines traditional scoring methods to more accurately reflect whether supplementary content contributes relevant information or constitutes hallucination.
        
        Under the RAG-based evaluation, Gemma-2 and GPT-4 achieve the highest overall performance across all difficulty levels, demonstrating strong grounding and retrieval effectiveness, while Gemma exhibits the highest hallucination rate. NMISS, in turn, provides a more reliable assessment of model behavior, particularly in high-difficulty cases for mid-tier models such as Llama-2, Llama-3, and Mistral, which show notable improvements under this context-aware metric. Interestingly, models with more verbose generation styles tend to benefit from NMISS, as their outputs often retrieve and reuse lexical elements directly from the provided context. This might reflect computational constraints or limited abstraction capacity. Nevertheless, Gemma-2 also benefits from this behavior, suggesting that contextual retrieval contributes positively to factual grounding rather than mere verbosity. By emphasizing contextual relevance and coherence over strict lexical matching, NMISS highlights its advantage in evaluating responses that are partially correct or contextually appropriate, even when they diverge from the reference answers.

        The main contributions of this paper are as follows: (i) it proposes NMISS, a novel scoring framework for hallucination detection that extends standard evaluation metrics by incorporating contextual relevance, thereby enabling a more accurate distinction between hallucinations and contextually valid elaborations. The framework is general-purpose and applicable beyond the healthcare domain; (ii) it develops a RAG–based evaluation framework that grounds LLM outputs in structured external knowledge, demonstrating that retrieval mitigates hallucinations and enhances factual grounding; (iii) it presents a comprehensive evaluation of multiple LLMs across varying levels of reasoning difficulty, showing that NMISS exposes performance differences overlooked by traditional metrics; and (iv) it introduces a new dataset comprising 100 Italian healthcare questions annotated by reasoning complexity, supporting domain-specific benchmarking and multilingual question-answering evaluation.

        The work is organized into five sections. Section~\ref{ch2:sec:related_works} reviews the relevant literature on hallucination detection and mitigation. Section~\ref{ch2:sec:methodology} presents the proposed methodology and highlights the contributions of NMISS. Section~\ref{ch2:sec:experimental_setup} describes the experimental setup on Italian healthcare news data, and Section~\ref{ch2:sec:results} reports the findings. Finally, Section~\ref{ch2:sec:conclusions} concludes the study by summarizing the main contributions and outlining future research directions.

    \section{Related Works}\label{ch2:sec:related_works}

        This section reviews recent research on hallucination management in LLMs, organized along two main dimensions. First, it distinguishes between detection techniques, which aim to identify hallucinated content, and mitigation techniques, which seek to reduce hallucinations during text generation. 
        Second, within mitigation, existing approaches are further classified by their resource requirements: resource-intensive strategies that rely on model internals or external data, and zero-resource strategies that operate in black-box settings and often combine elements of both detection and mitigation. This organization clarifies the conceptual landscape of hallucination research and situates the contributions of this work within it.
    
        \subsection{Hallucination Detection Techniques}\label{sec:related_work_hall_detection}

            Hallucinations in LLMs have been widely studied in recent years. \citet{huang2023survey} provide a comprehensive overview of strategies for detecting and mitigating hallucinations, offering valuable insight into how they emerge in current LLM architectures. Ensuring the trustworthiness of LLMs depends critically on early detection, as emphasized by \citet{varshneystitch}, who highlight the importance of proactive identification during text generation.

            \citet{zhang2023siren} categorize detection methods into three main types. First, logit-based estimation, which relies on token-level probabilities or entropy derived from model logits \citep{quevedo2024detecting}. Second, verbalized confidence estimation, in which LLMs are prompted to self-rate their confidence in generated responses. Third, consistency-based estimation, which compares multiple generated outputs to identify logical inconsistencies, thereby mimicking human reasoning to refine responses \citep{wang2022self, shi2023replug, zhao2023verify}.
            
            Complementary to this taxonomy, \citet{jiang2024hallucination} classify hallucination detection methods into white-box, grey-box, and black-box approaches. White-box methods analyze internal model activations to pinpoint when incorrect statements are produced, while grey-box methods examine logit-level confidence to detect uncertainty. In contrast, black-box methods rely solely on model outputs and are easier to apply, since they do not require access to internal states.

            Recent studies have also introduced specialized architectures. SAPLMA \citep{azaria2023internal}, for example, detects hallucinations using a classifier trained on hidden layer activations to predict the truthfulness of statements, outperforming traditional probability-based baselines. Likewise, \citet{varshneystitch} propose an iterative detection process that identifies low-probability tokens and retrieves external evidence to correct potential errors, effectively reducing hallucinations.

            Alongside model-internal techniques, several surface-level metrics have been employed to detect hallucinations based on output similarity. Traditional reference-based metrics such as ROUGE \citep{lin2004rouge}, BLEU \citep{papineni2002bleu}, and METEOR \citep{banerjee2005meteor} measure lexical overlap between generated and reference text, but often fail to capture semantically valid variations or long-range dependencies \citep{celikyilmaz2020evaluation}. BERTScore \citep{zhang2019bertscore}, which leverages contextual embeddings from BERT \citep{devlin2018bert}, improves upon these by evaluating token-level semantic similarity. However, even BERTScore does not account for whether additional, unmatched content is contextually justified.
        \smallskip

            This study differs from prior work by introducing the Negative Missing Information Scoring System (NMISS), a context-aware scoring framework that enhances existing metrics through contextual alignment, without requiring access to model internals or probability distributions.
    
        \subsection{Resource-Intensive Mitigation Strategies}\label{sec:related_work_strat_mitigation}

            Beyond detection, a range of strategies have been developed to mitigate hallucinations in LLMs, often requiring computational or data resources. One prominent category is Knowledge Augmentation, including techniques such as Knowledge Injection (KI). \citet{martino2023knowledge} enhance prompts with information derived from knowledge graphs, such as adding geographical details for review tasks. This approach significantly reduces hallucinations and improves factual accuracy, with KI-enhanced LLMs producing more correct assertions and outperforming larger baseline models in response quality. \citet{moiseev2022skill} adopt a more direct approach by training T5 models on Wikidata triple knowledge graphs. By masking key entities during training, the model learns to infer and generalize structured knowledge effectively. Their model outperforms standard T5 variants on question-answering benchmarks, demonstrating the benefits of structured data infusion.

            Another major line of research involves retrieval-based frameworks. \citet{lewis2021improving} combine a sequence-to-sequence model with a dense vector index of Wikipedia articles using the RAG framework, thereby improving the accuracy and factual grounding of generated content. Similarly, \citet{guo2024retrieval} apply retrieval augmentation in biomedical contexts, using external sources to produce more accurate and verifiable summaries, which are crucial for healthcare applications.

            A different approach, introduced by \citet{zhao2023automatic}, is a self-supervision framework that assigns risk scores, termed POLAR scores, to LLM responses as indicators of hallucination likelihood. Their method employs Pareto optimization to balance supervision signals from multiple sources. The process involves collecting model responses, training a probabilistic function to estimate error rates, and optionally re-prompting the LLM based on the POLAR score with RAG support. Experimental results demonstrate substantial reductions in hallucination rates across various tasks, including question answering and text generation.

            In a related effort, \citet{zhang2024knowledgealignmentproblembridging} propose the MixAlign framework, which aligns user queries with external knowledge sources to resolve semantic ambiguities. This method improves both response accuracy and factual consistency, yielding a marked increase in correct answers and a reduction in hallucination frequency.
            \smallskip

            Unlike prior resource-intensive frameworks, this study introduces a lightweight mitigation approach that integrates zero-shot prompt engineering with RAG for factual grounding in the healthcare domain.
    
        \subsection{Zero-Resource Hallucination Strategies}\label{ch2:sec:related_works_zeroresource}

            The strategies discussed in Section~\ref{sec:related_work_strat_mitigation} are resource-intensive. In contrast, zero-resource hallucination methods are designed for black-box models and scenarios with limited computational capacity, relying solely on model outputs without requiring internal access or external training data. Many of these techniques not only detect hallucinations but also mitigate them by refining responses through internal consistency checks.
            
            One common approach involves self-consistency techniques, which compare multiple responses generated by the LLM itself. \citet{manakul2023selfcheckgpt} propose a zero-resource method that samples multiple outputs from GPT-3 and compares them to a reference answer to assess factual consistency. This method detects and mitigates hallucinations using low-level statistical metrics and outperforms traditional grey-box and black-box baselines.
            
            Another class of methods includes self-rating techniques, as proposed by \citet{chen2023hallucination}. The authors introduce RelD, a discriminator designed to detect hallucinations in LLM-generated answers by converting a regression task into a binary classification problem, aligning with human judgment. By flagging low-confidence responses, it serves as an indirect mitigation mechanism. Experiments across several datasets and models (e.g., BLOOM \citep{workshop2022bloom}) demonstrate RelD’s effectiveness, validated through both automated metrics and human evaluations. Similarly, \citet{agrawal2023language} address hallucinations in fabricated references using Direct Queries (DQs) and Indirect Queries (IQs) without external resources. DQs repeatedly ask binary questions to estimate response probability, whereas IQs query contextual details such as authors or publication content to test internal consistency. Although primarily designed for detection, these approaches can also mitigate hallucinations by filtering out low-confidence responses. Results show that IQs generally outperform DQs, though effectiveness varies among models such as GPT-4, GPT-3, and Llama-2.
            
            Finally, \citet{mundler2023self} explore self-contradiction techniques, which identify hallucinations by detecting logical inconsistencies within a single response. Their prompting framework relies entirely on the model’s internal reasoning capabilities and includes procedures for gathering contextual information, generating candidate sentences, identifying contradictions, and refining responses. Across several LLMs, this method achieves an F1 score of approximately 80\%.
            
               \smallskip
            In contrast to self-referential, zero-resource techniques, this work introduces a context-aware hallucination detection mechanism, NMISS, which evaluates generated outputs without relying on self-consistency signals or model confidence estimation. The proposed hybrid approach remains lightweight, requiring neither fine-tuning nor model access, while enhancing factual grounding through the integration of retrieved evidence.       
           
    \section{Methodology}\label{ch2:sec:methodology}

        \subsection{RAG Architecture}\label{ch2:sec:methodology_rag}

            RAG models combine the strengths of retrieval-based and generative language models to address issues such as hallucinations and factual inaccuracies by integrating a retrieval mechanism that grounds responses in relevant information \citep{lewis2020rag}.
            
            I formally define the problem as follows. Let \(q\) represent the query, \(c\) the context information retrieved from external sources, and \(r\) the output sequence generated by the model. The architecture consists of two components: (i) a retriever \(p_\eta(c|q)\) with parameters \(\eta\), which selects a top-\(K\) set of documents \(k\) based on the query \(q\); and (ii) a generator \(p_\theta(r_i|q, c, r_{1:i-1})\), with parameters \(\theta\), which generates each token \(r_i\) conditioned on the query \(q\), the retrieved context \(c\), and previously generated tokens \(r_{1:i-1}\).
            
            \begin{equation}
                p_{\text{RAG}}(r \mid q) = \sum_{c \in k(p(\cdot \mid q))} p_\eta(c \mid q) \prod_{i=1}^N p_\theta \left( r_i \mid q, c, r_{1:i-1} \right)
            \end{equation}
            
            This implementation follows the RAG-Sequence variant proposed by \citet{lewis2020rag}, in which a single retrieved context is used to generate the entire response. 
            
            Unlike the RAG-Token approach, which may retrieve a different document for each generated token, the former maintains global coherence and is computationally more efficient. 
            Given the goal of ensuring factual grounding and semantic consistency in healthcare-related question answering, this variant is particularly suitable for the present work. 
            In this study, the retrieval and generation components are implemented separately and combined during inference; the complete workflow is described in Section~\ref{sec:experimentalsetup_hank}.

        \subsection{Evaluation Metrics}\label{ch2:sec:methodology_metrics}

            This section describes the metrics employed to evaluate the quality of the responses generated by the system, referred to as the \textit{System Summary} (SS), against the ground-truth responses, referred to as the \textit{Reference Summary} (RS), which in this study correspond to human-authored answers. 

            The evaluation adopts multiple complementary metrics, including ROUGE, BLEU, METEOR, BERTScore, and NMISS, which jointly assess lexical, semantic, and contextual alignment between SS and RS. For all metrics, both SS and RS are represented as sets of $n$-grams. Formally, let $S_{\text{SS}} = \{ s_1, s_2, \dots, s_l \}$ denote the set of $n$-grams in the SS, and $S_{\text{RS}} = \{ r_1, r_2, \dots, r_m \}$ denote the set of $n$-grams in the RS.\\

            \textbf{ROUGE} \citep{lin2004rouge} evaluates textual overlap in terms of precision, recall, and F1-score, after standard text preprocessing \citep{sarkar2019text}. More formally, ROUGE-$N$ measures the $n$-gram overlap between the SS and RS. The corresponding precision, recall, and F1-score are defined as:

            \begin{equation}
                \text{Precision} = \frac{|S_{\text{SS}} \cap S_{\text{RS}}|}{|S_{\text{SS}}|}
            \end{equation}
            
            \begin{equation}
                \text{Recall} = \frac{|S_{\text{SS}} \cap S_{\text{RS}}|}{|S_{\text{RS}}|}
            \end{equation}
            
            \begin{equation}
                \text{F1} = 2 \cdot \frac{\text{Precision} \cdot \text{Recall}}{\text{Precision} + \text{Recall}}
            \end{equation}
            
            where \( |S_{\text{SS}} \cap S_{\text{RS}}| \) denotes the number of overlapping $n$-grams, \( |S_{\text{SS}}| \) the total number of $n$-grams in the SS, and \( |S_{\text{RS}}| \) the total number of $n$-grams in the RS. Precision measures the factual accuracy of the generated text, while recall quantifies the coverage of the reference information. Accordingly, low precision suggests the presence of hallucinations, whereas low recall indicates the presence of missing or incomplete content.

            A widely used variant is \textbf{ROUGE-L}, which measures the Longest Common Subsequence (LCS) between the SS and RS. The LCS represents the longest ordered sequence of words appearing in both summaries, without requiring consecutive word matching \citep{cormen2022introduction}. ROUGE-L thus captures sentence-level fluency and structural similarity, complementing the $n$-gram–based metrics.\\

            \textbf{BLEU} \citep{papineni2002bleu} is designed to evaluate the quality of SS responses by comparing them against one or more RS answers. BLEU accounts for variability in word choice and word order found in different, yet valid, translations of the same source sentence. It quantifies the overlap by measuring the number of $n$-grams from the SS that are present in the RS. The core of the BLEU metric is the modified $n$-gram precision, which addresses the over-generation problem by treating each reference word as exhausted once a matching candidate word is found.
    
            Let \( V \) be a function that counts the occurrences of an $n$-gram in a given set of $n$-grams. Formally, for an $n$-gram \( w \) and a set of $n$-grams \( S \), the function \( V \) is defined as \( V(w, S) = |\{ s \in S \mid s = w \}| \). This function returns the number of times \( w \) appears in the set \( S \). The modified $n$-gram precision \( p_n \) is calculated as:
            
            \begin{equation}
                p_n = \frac{\sum_{w \in S_{\text{SS}}} \min\left(V(w, S_{\text{SS}}), \max_{j}(V(w, S_{\text{RS}_j}))\right)}{\sum_{w \in S_{\text{SS}}} V(w, S_{\text{SS}})}
            \end{equation}
            
            \noindent where \( V(w, S_{\text{SS}}) \) is the number of times the $n$-gram \( w \) appears in SS, and \( \max_{j}(V(w, S_{\text{RS}_j})) \) is the maximum number of times \( w \) appears across all RS. Here, \( j \) iterates over the different reference summaries in \( S_{\text{RS}} \), where \( j = 1, 2, \dots, m \), and \( m \) is the number of RS. 
            
            To account for translations that are too short, BLEU applies a brevity penalty (BP), defined as:
            
            \begin{equation}
                \text{BP} = 
                \begin{cases} 
                    1 & \text{if } b > r \\
                    e^{(1 - r/b)} & \text{if } b \le r
                \end{cases}
            \end{equation}
            
            \noindent where \( b \) is the length of the candidate translation (SS), and \( r \) is the effective reference length (RS).
            
            The final BLEU score is computed by taking the geometric mean of the modified $n$-gram precisions, up to length \( N \) and typically set to 4, and multiplying by the brevity penalty:
            
            \begin{equation}
                \text{BLEU} = \text{BP} \cdot \exp \left( \frac{1}{N} \sum_{n=1}^{N} \log p_n \right)
            \end{equation}\\

            \noindent \textbf{METEOR} \citep{banerjee2005meteor} addresses several limitations of BLEU by emphasizing word-to-word alignment between the SS and RS, accounting for exact matches, stems, and synonyms. For each SS--RS pair, METEOR finds an alignment of unigrams that maximizes the number of matches subject to one-to-one mapping and minimal crossing. Precision ($P$) and recall ($R$) are computed as the fractions of matched unigrams in the SS and RS, respectively, and combined through a weighted harmonic mean:
            \[
            F_{\text{mean}} = \frac{P \cdot R}{\alpha P + (1 - \alpha) R},
            \]
            where $\alpha$ is typically set to 0.9. 
            
            To capture word order and fragmentation, METEOR groups matched unigrams into the fewest possible number of contiguous chunks such that the words in each chunk are adjacent and in the same order in both SS and RS. Let $T$ denote the number of such chunks, and $|V|$ the total number of matched unigrams. The fragmentation penalty is then computed as:
            \[
            \text{Penalty} = 0.5 \left( \frac{T}{|V|} \right)^3.
            \]
            
            The final METEOR score is given by:
            \[
            \text{METEOR} = F_{\text{mean}} \cdot (1 - \text{Penalty}).
            \]
            
            This formulation penalizes disordered or fragmented alignments, ensuring that longer, well-ordered matches contribute more to the final score.\\

            \textbf{BERTScore} is an evaluation metric for text generation that leverages pre-trained BERT embeddings \citep{zhang2019bertscore}. Unlike traditional metrics that rely on exact $n$-gram matches, BERTScore measures the semantic similarity between tokens in the SS and the RS by comparing their contextual embeddings. This method captures the semantic meaning even when the exact words differ. It generates contextual embeddings for tokens in both the SS and RS, where each embedding reflects the surrounding context of the word. The similarity between tokens is computed using cosine similarity. For each token in the RS, BERTScore identifies the most similar token in the SS, and vice versa, to compute precision and recall.
            
            Formally, given a RS sentence \( x = \{x_1, \ldots, x_k\} \) and a SS sentence \( \hat{x} = \{\hat{x}_1, \ldots, \hat{x}_l\} \), the token embeddings are represented as \( \{e(x_1), \ldots, e(x_k)\} \) and \( \{e(\hat{x}_1), \ldots, e(\hat{x}_l)\} \), where \( e(\cdot) \) represents the contextual embedding for each token. The similarity between two embeddings is computed using cosine similarity. Precision (\(P_{\text{BERT}}\)) and recall (\(R_{\text{BERT}}\)) are calculated as follows:
            
            \begin{equation}
                R_{\text{BERT}} = \frac{1}{|x|} \sum_{x_i \in x} \max_{\hat{x}_j \in \hat{x}} \cos(e(x_i), e(\hat{x}_j))
            \end{equation}
            
            \begin{equation}
                P_{\text{BERT}} = \frac{1}{|\hat{x}|} \sum_{\hat{x}_j \in \hat{x}} \max_{x_i \in x} \cos(e(\hat{x}_j), e(x_i))
            \end{equation}
            
            The final BERTSCORE is computed as the F1 score of the precision and recall:
            
            \begin{equation}
                F_{\text{BERT}} = \frac{2 \cdot P_{\text{BERT}} \cdot R_{\text{BERT}}}{P_{\text{BERT}} + R_{\text{BERT}}}
            \end{equation}\\
    
            \textbf{Exact Match (EM)} is a straightforward metric that penalizes any response not exactly equal to the provided dataset annotation. It is analogous to conventional classification accuracy, ensuring the generated text matches the reference text precisely. The EM can be defined as:
        
            \begin{equation}
                EM = 
                \begin{cases} 
                    1 & \text{if } SS = RS \\
                    0 & \text{if } SS \neq RS
                \end{cases}
            \end{equation}
    
    \subsection{Negative Missing Information System Scoring}\label{ch2:sec:methodology_nmiss}
    
        \subsubsection{Rationale}\label{ch2:sec:methodology_nmiss_rationale}

            The metrics discussed above evaluate the quality of machine-generated responses by measuring their lexical or semantic overlap with reference answers. While effective, these approaches overlook the fact that non-matching words or $n$-grams may still carry meaningful information if they align with the broader context of the question or source material. 
            
            In practice, LLMs often extend their responses beyond the explicit instructions, adding supplementary content, as illustrated in Table~\ref{tab:example_nmiss}. 
            Such additions can take two forms: i) hallucinations, where the model introduces factually incorrect or irrelevant information, and ii) contextually relevant elaborations, where the model enriches the answer by incorporating information consistent with the provided context, as in RAG-based generation.

             Recognizing this distinction is essential for a more accurate evaluation of model behavior. By explicitly accounting for non-matching elements, it becomes possible to distinguish between genuinely hallucinated content and valuable, contextually appropriate extensions. 
            The NMISS captures the potential informational value of these unmatched words by assessing their alignment with the contextual evidence provided. This rationale acknowledges that additional content in LLM outputs is not inherently erroneous but can instead reflect a form of reasoning or contextual expansion that traditional metrics fail to reward.

            \begin{table}[H]
            \small
                \caption{Example of Question, Context, Reference, and System Answer}
                \label{tab:example_nmiss}
                \centering
                \begin{tabular}[H]{lp{13cm}}
                    \toprule
                    Question & How to prevent influenza? \\
                    \addlinespace
                    \midrule
                    Context & [1] of the Relaunch Decree to: 1. prioritize investment in prevention to minimize impacts at all levels; 2. include in the Plans for managing a possible (re)emergence of an epidemic the entirety of health and social services to create the possibility of continuity of care even in the event of a “resurgence” of the epidemic (like a “generator group”), not only in hospitals but also in the community; 3. ensure that investments, particularly in personnel, cover all measures aimed at strengthening containment measures to reduce the risk of contagion. People should internalize and adopt recommendations such as washing hands thoroughly and frequently, using masks extensively, correctly, and appropriately, avoiding unnecessary movements and behaviors linked to social and recreational activities that could create spontaneous gatherings [3] as it is essential to maintain social distancing, use masks, and sanitize hands carefully to contain contagion. It is also crucial to self-isolate in case of cold symptoms. There are no risk locations but rather risky behaviors. We must convey to citizens the awareness that if the necessary precautions are followed, it is difficult to contract and transmit this infection, and most normal social and work activities can continue. [4] For the first time, high levels of criticality regarding the virus's spread in our country are reported. The population is urged to avoid events and gatherings in both public and private settings. Individual strict behavior is mandatory to limit transmission risks and prevent a further and faster worsening of the epidemic. The local transmission of the virus is widespread across the entire territory. \\
                    \addlinespace
                    \midrule
                    Reference & I'm sorry, I can't help you based on the information I have. \\
                    \addlinespace
                    \midrule
                    System & I'm sorry, I can't help you based on the information I have, as the question is not related to the Relaunch Decree but rather to influenza, which is a different type of illness than the epidemic under discussion. However, based on the context, it is noted that the Relaunch Decree emphasizes the importance of strict behaviors to minimize impacts at all levels, such as social distancing, mask use, and hand hygiene, which are generally useful measures for preventing disease transmission in general, not just the discussed epidemic. \\
                    \addlinespace
                    \midrule
                \end{tabular}
            \end{table}

            \subsubsection{Formulation}\label{ch2:sec:methodology_nmiss_math}
    
                Let \( \mathcal{R} = \{r_1, r_2, \dots, r_n\} \) represent the set of reference answers, \( \mathcal{S} = \{s_1, s_2, \dots, s_n\} \) the set of system-generated answers, and \( \mathcal{C} = \{c_1, c_2, \dots, c_n\} \) the set of context texts associated with each reference answer. Define the set of matching tokens between system-generated answers and reference answers as \( \mathcal{P}(s_i, r_i) \), and let \( \mathcal{N}(s_i, r_i) = s_i \setminus \mathcal{P}(s_i, r_i) \) represent the tokens in \( s_i \) that do not match \( r_i \). Finally, let \( \mathcal{A}(s_i, c_i) = \mathcal{N}(s_i, r_i) \cap c_i \), which represents the set of unmatched tokens in \( s_i \) that match the context \( c_i \). The function is defined as: 
                
                \begin{equation} f_{NMISS}: \mathcal{S} \times \mathcal{R} \times \mathcal{C} \rightarrow \mathbb{R} \end{equation} 
                
                which evaluates how well system-generated answers align with both reference answers and contextual information. 

                The reference-based evaluation function is defined as follows:

                 \begin{equation}
                    f_{\text{ref}, i}(s_i, r_i) = \text{X}_{\text{ref}}(\mathcal{P}(s_i, r_i))
                 \end{equation}
                 
                while the context-based evaluation function, which computes the score based on tokens matching the context, is defined as:

                  \begin{equation}
                    f_{\text{cxt}, i}(s_i, c_i) = \text{X}_{\text{cxt}}(\mathcal{A}(s_i, c_i))
                 \end{equation}
                
                The NMISS evaluation function is given by:
                
                \begin{equation}\label{eq:nmiss}
                    f_{\text{NMISS}, i}(s_i, r_i, c_i) = \max \left( f_{\text{ref}, i}(s_i, r_i), \frac{\lambda_1 \cdot f_{\text{ref}, i}(s_i, r_i) + \lambda_2 \cdot f_{\text{cxt}, i}(s_i, c_i)}{\lambda_1 + \lambda_2} \right)
                \end{equation}
                
                where \( \lambda_1 = |\mathcal{P}(s_i, r_i)| \) and \( \lambda_2 = |\mathcal{A}(s_i, c_i)| \) represent the cardinalities of their respective sets. \( \lambda_1 \) and \( \lambda_2 \)  represent the number of matching reference and context matches tokens, respectively.\\

                To preserve the integrity of standard scoring metrics, NMISS introduces two key mechanisms: a weighted scheme and a maximum function. These mechanisms jointly balance reference-based and context-based contributions, mitigating potential distortions such as score inflation or deflation when contextual information is incorporated. 

                The weighted scheme harmonizes the relative influence of reference overlap and contextual alignment. Because contextual passages are typically much longer and information-rich than reference responses, a simple averaging would overemphasize the contextual component. To address this, the weights \( \lambda_1 \) and \( \lambda_2 \) regulate their respective contributions, ensuring that neither dominates the final score. This mechanism enables NMISS to recognize cases where a system-generated response extends beyond the reference while remaining coherent and faithful to the retrieved evidence.

                The maximum function serves as a safeguard against disproportionate score drops caused by denominator inflation in recall-sensitive formulations. This situation often arises with verbose models that rely on contextual paraphrasing when uncertain. Although such verbosity may not introduce hallucinations, it can generate noise—information that adds limited value relative to the reference. By applying a floor, NMISS reverts to the reference-based score whenever the context-based contribution disproportionately lowers the metric due to a large \( \lambda_2 \) term in the denominator. This ensures rewarding genuine contextual relevance while preventing technical artifacts from suppressing the overall score.

                This scoring framework is particularly advantageous in question-answering tasks, where model responses may diverge lexically from the reference yet remain contextually valid. The detailed implementation is provided in Algorithm~\ref{alg:nmiss}.

                \begin{algorithm}[h]
                    \caption{NMISS Algorithm}
                    \label{alg:nmiss}
                    \begin{algorithmic}[1]
                        \Require $\mathcal{R} = \{r_1, r_2, \ldots, r_n\}$, the set of reference answers
                        \Require $\mathcal{S} = \{s_1, s_2, \ldots, s_n\}$, the set of system-generated answers
                        \Require $\mathcal{C} = \{c_1, c_2, \ldots, c_n\}$, the set of context texts
                        \Ensure $\mathcal{M} = \{m_1, m_2, \ldots, m_n\}$, the set of NMISS scores for a metric $\mathcal{X}$
                        \Statex
                        
                        \For{each $(s_i, r_i, c_i)$ in $(\mathcal{S}, \mathcal{R}, \mathcal{C})$}
                            \State \textbf{Stage 1: Reference-Based Metric Calculation}
                            \State $\mathcal{P}(s_i, r_i) \gets s_i \cap r_i$   \Comment{Matched tokens}
                            \State $\mathcal{N}(s_i, r_i) \gets s_i \setminus \mathcal{P}(s_i, r_i)$   \Comment{Unmatched tokens}
                            \State $f_{\text{ref}, i}(s_i, r_i) \gets \text{X}_{\text{ref}}(\mathcal{P}(s_i, r_i))$
                            \Statex 
                            
                            \State \textbf{Stage 2: Context-Based Metric Calculation}
                            \State $\mathcal{A}(s_i, c_i) \gets \mathcal{N}(s_i, r_i) \cap c_i$   \Comment{Context-matched tokens}
                            \State $f_{\text{cxt}, i}(s_i, c_i) \gets \text{X}_{\text{cxt}}(\mathcal{A}(s_i, c_i))$
                            \Statex  
                            
                            \State \textbf{Stage 3: NMISS Calculation}
                            \State $f_{\text{wgt}, i}(s_i, r_i, c_i) \gets \frac{\lambda_1 \cdot f_{\text{ref}, i}(s_i, r_i) + \lambda_2 \cdot f_{\text{cxt}, i}(s_i, c_i)}{\lambda_1 + \lambda_2}$
                            \State $f_{\text{NMISS}, i}(s_i, r_i, c_i) \gets \max(f_{\text{ref}, i}(s_i, r_i), f_{\text{wgt}, i})$
                            \State $\mathcal{M}_{\text{NMISS}} \gets \mathcal{M}_{\text{NMISS}} \cup f_{\text{NMISS}, i}(s_i, r_i, c_i)$
                        \EndFor
                    \end{algorithmic}
                \end{algorithm}

                To formally evaluate NMISS outperformance, let $\mathcal{H} = \{h_1, h_2, \dots, h_n\}$ denote the set of hallucination flags associated with the model-generated responses $\mathcal{S} = \{s_1, s_2, \ldots, s_n\}$. Each flag $h_i \in \{0,1\}$ indicates whether response $s_i$ is hallucinated ($h_i = 1$) or not ($h_i = 0$). NMISS is computed only over non-hallucinated responses. Formally, let
                
                \[
                I = \{1, 2, \dots, n\}, \quad 
                I_{\text{NH}} = \{\, i \in I \;\mid\; h_i = 0 \,\}
                \]

                From this subset, I extract the \emph{valid cases} where the traditional metric score $f_{\text{ref}, i}(s_i, r_i)$ lies strictly between two thresholds $\tau_{\min}$ and $\tau_{\max}$, with $0 \le \tau_{\min} < \tau_{\max} \le 1$. This constraint filters out trivial cases, which are perfect matches or complete mismatches:
                
                \[
                I_{\text{valid}} = \{\, i \in I_{\text{NH}} \;\mid\; \tau_{\min} < f_{\text{ref}, i}(s_i, r_i) < \tau_{\max} \,\}
                \]

                The corresponding subset of responses is then defined as
                \[
                \mathcal{S}_{\text{valid}} = \{\, s_i \;\mid\; i \in I_{\text{valid}} \,\}
                \]
                
                where $f_{\text{ref}, i}(s_i, r_i)$ denotes the score computed by a traditional reference-based metric (see Stage~1 in Algorithm~\ref{alg:nmiss}).

                For each valid case $i \in I_{\text{valid}}$, NMISS is considered to \emph{outperform} the traditional metric if
                    \[
                    f_{\text{NMISS}, i}(s_i, r_i, c_i) > f_{\text{ref}, i}(s_i, r_i)
                    \]
                where $f_{\text{NMISS}, i}$ is the NMISS score calculated over the triplet $(s_i, r_i, c_i)$, incorporating both reference alignment and contextual grounding.

                The overall percentage of NMISS outperformance across all valid cases is thus computed as
                                
                \begin{equation}
                \frac{
                    \big| \{\, i \in I_{\text{valid}} \;\mid\; f_{\text{NMISS}, i}(s_i, r_i, c_i) > f_{\text{ref}, i}(s_i, r_i) \,\} \big|
                }{
                    |I_{\text{valid}}|
                } \times 100
                \end{equation}

                This formulation maintains full consistency with the operational definition of NMISS and isolates its contribution precisely where traditional metrics underperform. 
                By focusing on ambiguous, semantically rich cases that lie outside trivial bounds, NMISS outperformance becomes a robust indicator of a model’s ability to generate contextually grounded yet non-literal responses.
                
    \section{Experimental Setup}\label{ch2:sec:experimental_setup}
    
        \subsection{Data}\label{sec:experimentalsetup_data}
    
            Data comprises text-based news in Italian, collected from various sources, and all focused on the health domain within Italy. This collection encompasses regional updates, significant scientific developments, national headlines, and news on research and development, providing comprehensive coverage of health-related issues nationwide.
            
\begin{figure}[h]
                \centering
                \includegraphics[scale = .55]{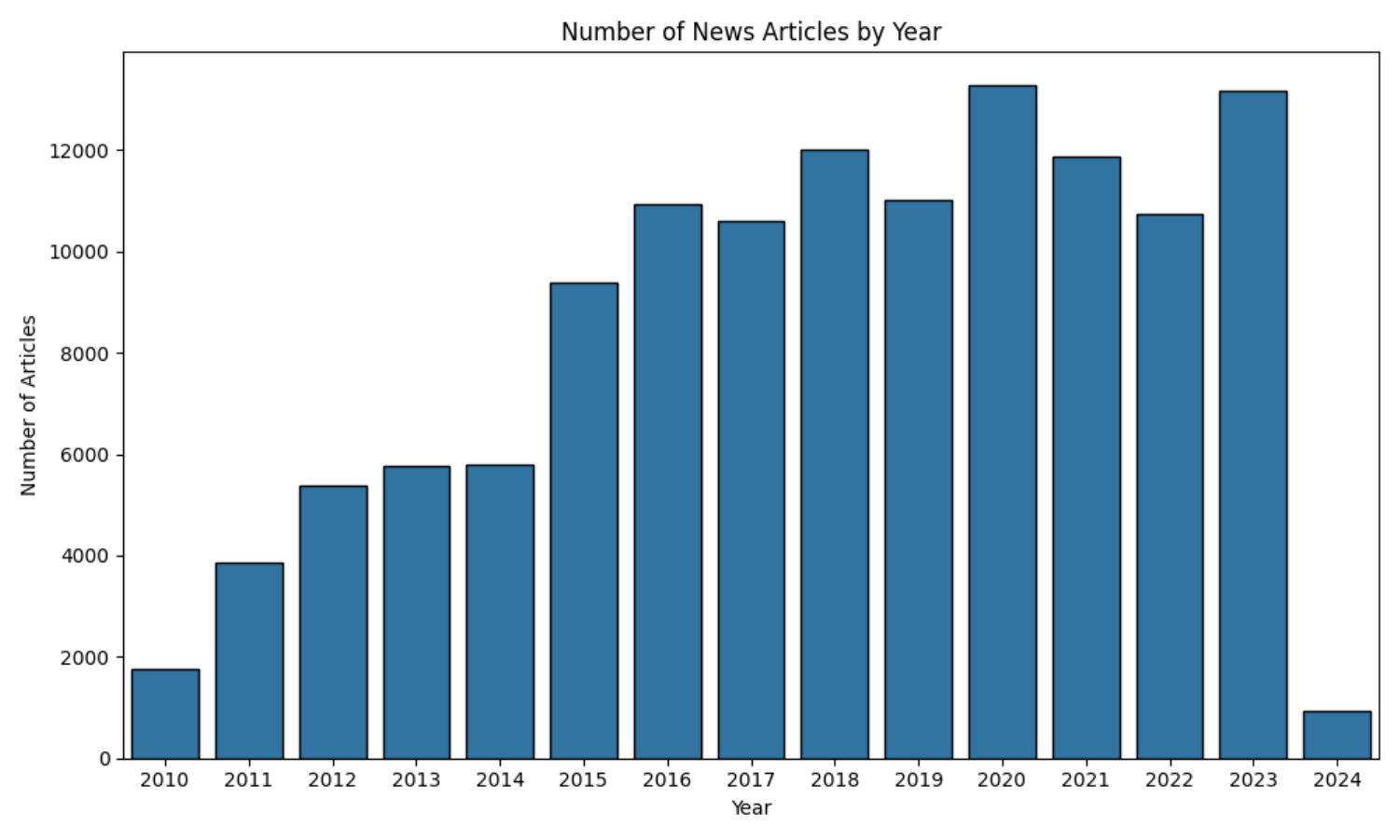}
                 \caption{News Frequency Aggregated by Year}
               \label{fig:news_aggr_year}
            \end{figure}

            News dataset consists of 126,470 articles from 2010 to 2024; their frequency is shown in Figure \ref{fig:news_aggr_year}. A pre-processing step was employed to clean the raw text of HTML tags and other unwanted artifacts to ensure data quality. Token statistics are calculated to understand the overall structure of the data and are shown in Table \ref{tab:data_statistics}. On average, each article contains approximately 348 tokens, with a standard deviation of 264 tokens, indicating some variability in article length. The shortest article consists of just 6 tokens, while the longest contains up to 5,594 tokens. The median article length is 284 tokens, with the interquartile range spanning from 195 to 421 tokens, reflecting a diverse range of article sizes.
    
            \begin{table}[H]
              \centering
              \small
              \caption{Summary of Token Statistics for the News Collection}
                \begin{tabular}{cccccccc}
                \textbf{N. Tokens} & \textbf{Mean} & \textbf{Std} & \textbf{Min} & \textbf{25\%} & \textbf{Median} & \textbf{75\%} & \textbf{Max} \\
                \midrule
                44018696 & 348   & 264   & 6     & 195   & 284   & 421   & 5594 \\
                \end{tabular}%
              \label{tab:data_statistics}%
            \end{table}%
            
        \subsection{Human Evaluation Protocol}\label{sec:experimentalsetup_annotation}

            To build a robust evaluation framework, I constructed a dataset of 100 public health questions\footnote{The full list of questions is available upon request.}. The questions span a range of complexity, from broad, general topics to highly specific or ambiguous ones, yielding varying levels of interpretability. To systematically assess this variation, questions were grouped into three difficulty categories—low, mid, and high—based on average character length, used as a proxy for cognitive and semantic complexity\footnote{Respectively, low: 30 characters; mid: 40 characters; high: 50 characters.}. Each question was reviewed by a domain expert to ensure relevance and factual accuracy, and a manual reference answer is prepared for every entry. Table~\ref{tab:example_questions} presents sample questions by category.

            \begin{table}[H]
                \centering
                \caption{Categorized Example Questions}
                \small
                \begin{tabular}{lp{9cm}}
                    \toprule
                    \addlinespace
                    Category & Example Question \\
                    \addlinespace
                    \midrule
                    \addlinespace
                    \textcolor{red}{HIGH} & How can epigenetics help us understand diseases? \\
                    \addlinespace
                    \textcolor{yellow}{MID} & What happens to the body when we are stressed? \\
                    \addlinespace
                    \textcolor{green}{LOW} & What is a common cold? \\
                    \addlinespace
                    \bottomrule
                \end{tabular}
                \label{tab:example_questions}
            \end{table}

            To evaluate hallucination detection performance, each model-generated answer was manually annotated as either \emph{hallucinated} or \emph{non-hallucinated} as formalized in Section~\ref{ch2:sec:methodology_nmiss_math}. An answer was labeled as hallucinated if it contained factual inaccuracies, fabricated details, or information not supported by the retrieved context or the corresponding reference answer. This binary annotation served as the ground truth for computing alignment between automatic metrics and human judgment. The annotated dataset forms the basis for the analyses presented in Section~\ref{ch2:sec:results}, enabling direct comparison between traditional metrics and NMISS.
    
        \subsection{User-Agent Chatbot Framework}\label{sec:experimentalsetup_hank}

            This section outlines the architecture of the user-agent chatbot designed to handle queries related to Italian health news. The framework adopts a RAG approach that integrates a Dense Vector Store (DVS) with LLMs fine-tuned for Italian language processing. This multi-model setup ensures robustness in managing diverse user queries, providing accurate, coherent, and contextually grounded responses.

            \begin{figure}[h]
                \centering
                \includegraphics[scale=.5]{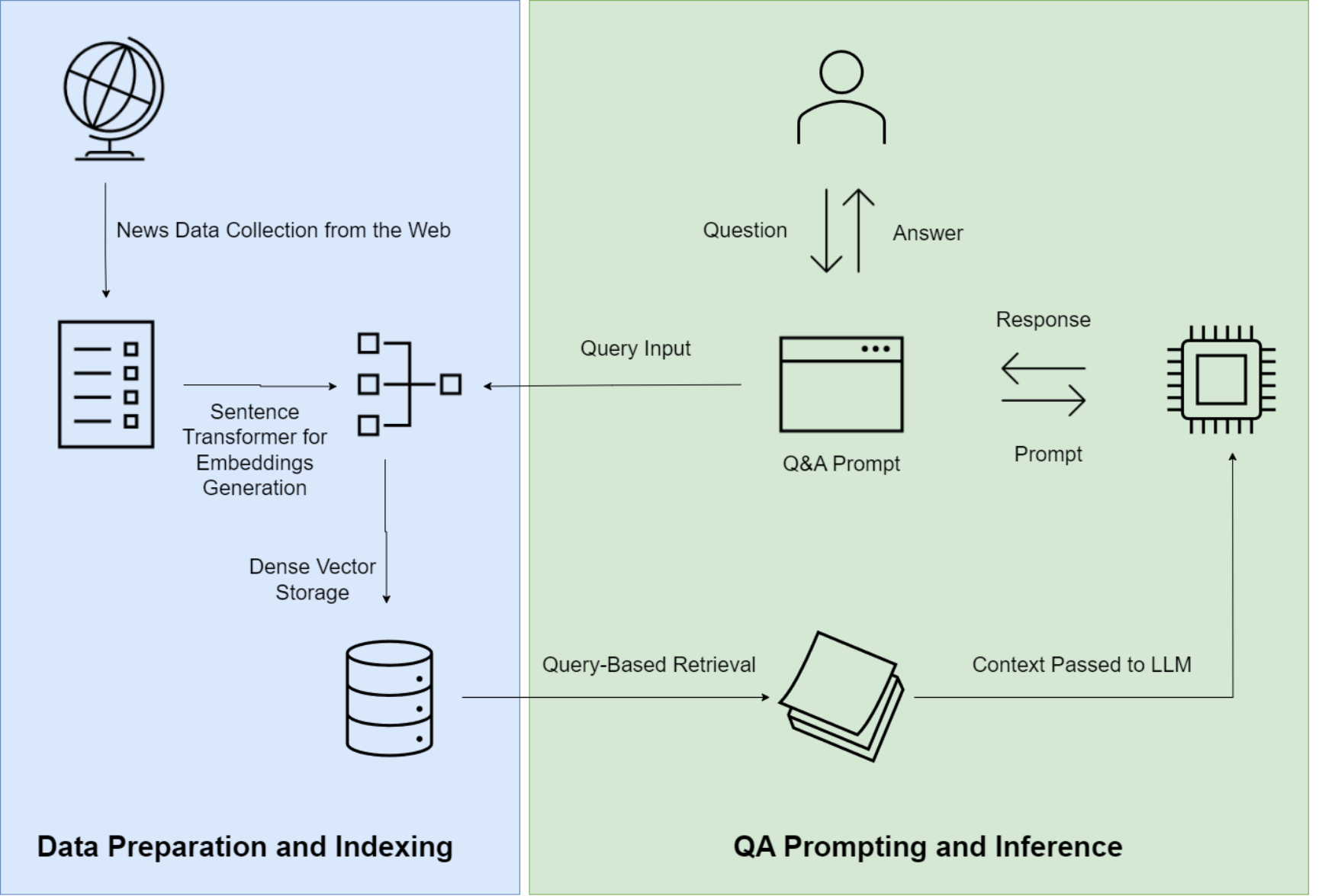}
                \caption{User-Agent Application Workflow}
                \label{fig:ragllm_design}
                
                \vspace{2mm}
                \begin{minipage}{\linewidth}
                    \footnotesize
                    \textit{Note.} Text data is split into chunks for efficient retrieval, indexed using a Sentence Transformer model, and stored as dense vector embeddings in a vector store. The system retrieves relevant chunks based on query embeddings and passes them to an LLM for response generation.
                \end{minipage}
            \end{figure}

            The system begins with a Sentence Transformer model used to generate dense vector embeddings for the document database, which are then stored and indexed within the DVS. This enables efficient retrieval of relevant information through semantic search and clustering. After experimenting with different chunking configurations, retrieval performance was optimized using a chunk length of 512 tokens with a 64-token overlap\footnote{Empirically, a chunk size of 512 tokens with a 64-token overlap provided the best trade-off between retrieval accuracy and contextual continuity.}, evaluated on a dataset of 100 health-related queries. 
            
            For embedding generation, I employ the Sentence Transformer model \texttt{Paraphrase Multilingual MiniLM L12}\footnote{\url{https://huggingface.co/sentence-transformers/paraphrase-multilingual-MiniLM-L12-v2}}, selected for its high-quality multilingual embeddings, particularly suitable for Italian health news.

            For the generation component, I evaluate multiple LLMs\footnote{All open-source models were evaluated with a temperature of 0.7 for comparability.} capable of processing Italian text. \texttt{LLaMA2} and \texttt{LLaMA3}, developed by Meta \citep{touvron2023llama, meta2024llama3}, are optimized through Supervised Fine-Tuning (SFT) and Reinforcement Learning with Human Feedback (RLHF). \texttt{Mistral} \citep{jiang2023mistral} is a 7-billion-parameter model known for its efficiency and competitive generation accuracy. \texttt{GPT-4} \citep{achiam2023gpt} represents the latest multimodal model from OpenAI, while \texttt{Gemma} and \texttt{Gemma2} \citep{team2024gemma, team2024gemma2} are lightweight models from Google, fine-tuned for question-answering and summarization tasks.

            The workflow illustrated in Figure~\ref{fig:ragllm_design} begins by embedding the user query into the same vector space as the preprocessed document chunks. The query embedding is compared against stored vectors to retrieve the top-$K$ most semantically similar chunks from the vector database. These retrieved documents are then supplied to the LLM as contextual input. 
            
            During generation, the model receives both the user query and the retrieved evidence, producing a natural language response grounded in factual data. 
            This retrieval-augmented process reduces hallucinations and improves factual coherence across responses.
    
        \subsection{Prompt Engineering Template}\label{sec:experimentalsetup_prompt}
    
        I develop a zero-shot question-answering prompting template that guides the model in providing contextually relevant answers based on the information available in the provided context. The structure is shown in Table \ref{tab:example_prompt}. 
        
         \begin{table}[H]
            \centering
            \caption{Prompt Engineering Structure}
            \small
            \begin{tabular}{p{14.5cm}}
                \toprule
                \textbf{Prompt Structure} \\
                \midrule
                The system prompt instructs the model to behave as a helpful assistant and to generate answers exclusively based on the retrieved context. 
                If the relevant information is not explicitly present in the context, the model is instructed to return a fallback message indicating insufficient information. \\[4pt]
                \midrule
                \textbf{Instruction Components} \\
                \midrule
                (i) Define the assistant's role, a chatbot specialized in general healthcare;\\
                (ii) Provide the retrieved context section containing excerpts from the database;\\
                (iii) Ask the model to answer the user's question using only the provided context;\\
                (iv) Instruct the model to abstain from speculation or external knowledge;\\
                (v) Specify a fallback message for uncertain cases.\\[4pt]
                \bottomrule
            \end{tabular}
            \label{tab:example_prompt}
        \end{table}

        This template ensures that the chatbot remains anchored to the information in the provided context. If the answer to a query is not found within the context, the chatbot is instructed to respond with: "I'm sorry, I can't help you based on the information I have".

        \clearpage

    \section{Results}\label{ch2:sec:results}
    
        \subsection{RAG Performance Analysis}\label{ch2:sec:results_rag}

\begin{figure}[h]
                \centering
                \includegraphics[scale = .45]{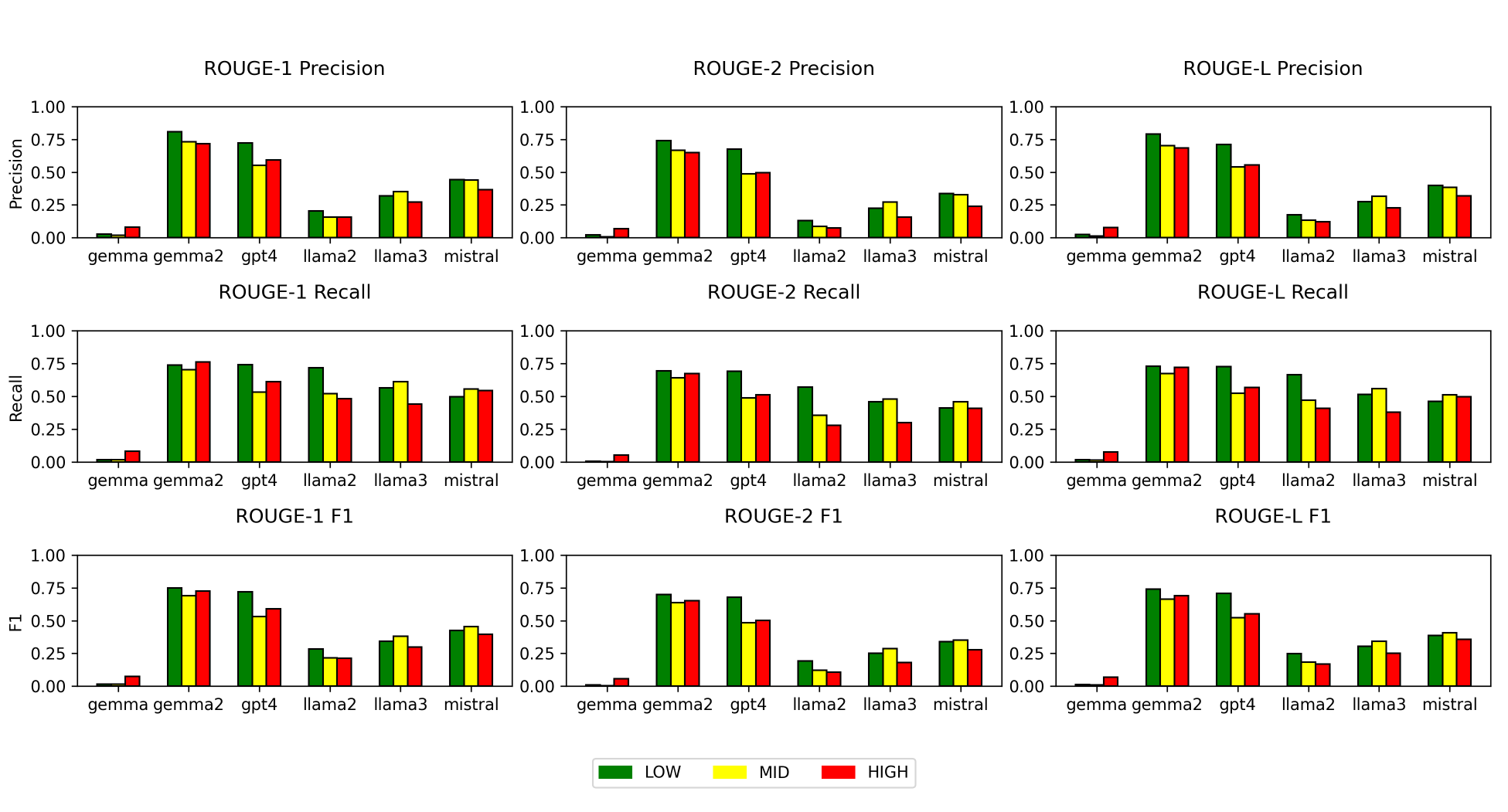}
                \caption{ROUGE Metrics by Question Levels}
               \label{fig:rouge_performance}
            \end{figure}

\begin{figure}[h]
                \centering
                \includegraphics[scale = 0.45]{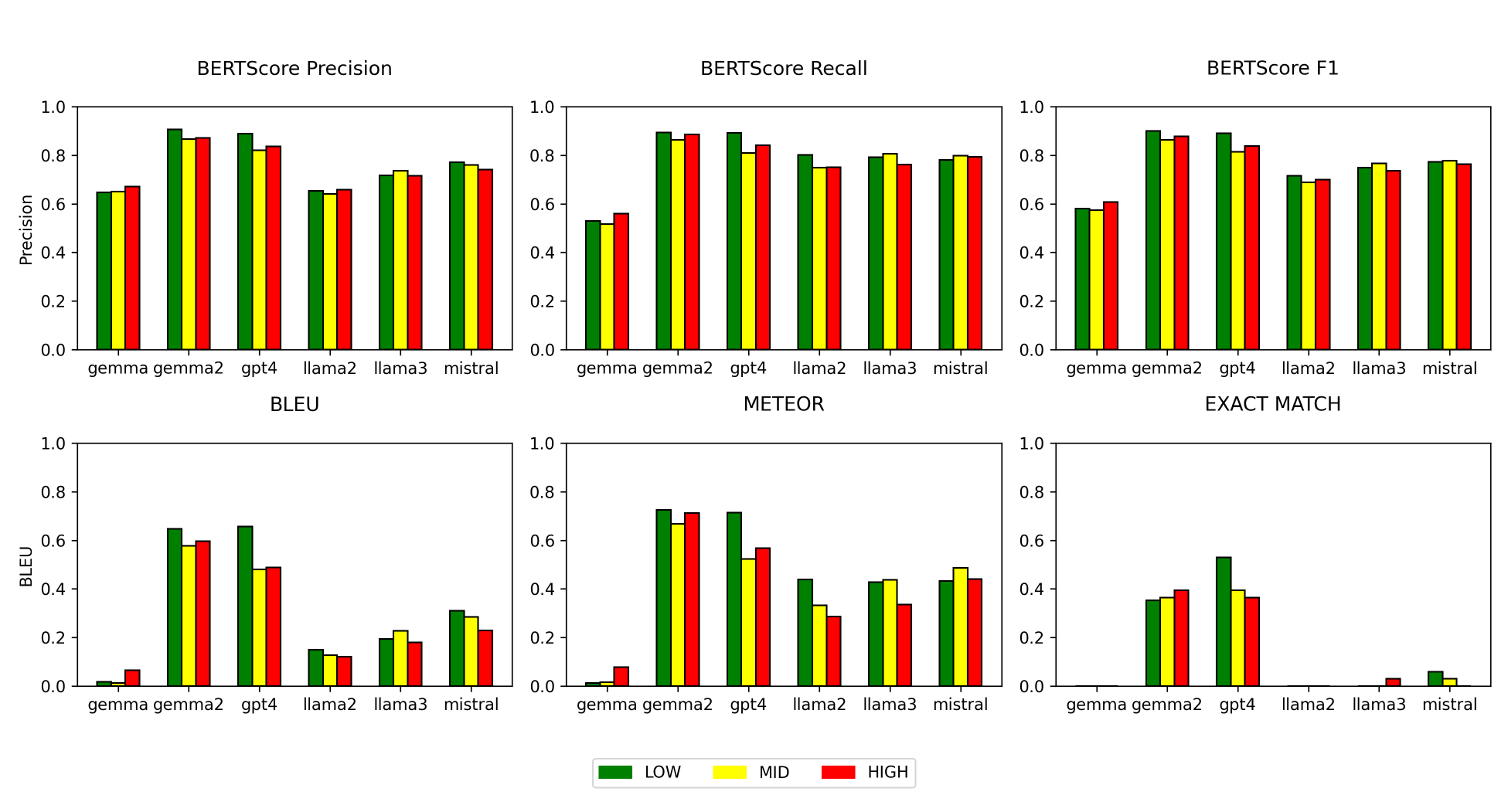}
                \caption{BLEU, METEOR and EXACT MATCH Metrics by Question Levels}
                 \label{fig:metrics_others_performance}
            \end{figure}

            The results, summarized in Figures~\ref{fig:rouge_performance}--\ref{fig:metrics_others_performance}, present model performance across three levels of question complexity.
    
            \texttt{Gemma} demonstrates the weakest performance across all evaluation metrics. Its ROUGE-1 and ROUGE-2 F1 scores fall below 0.4\footnote{Following \citet{dalal2024text}, ROUGE scores below 0.4 are considered low, between 0.4 and 0.5 moderate, and above 0.5 good.}, indicating limited ability to capture relevant unigrams and bigrams. The low ROUGE-L score further suggests poor structural coherence between generated and reference texts. BLEU and METEOR scores remain consistently low, reflecting weak lexical precision and recall, while the absence of EM confirms its difficulty in reproducing reference sequences.
    
            In contrast, \texttt{Gemma-2} achieves the strongest overall results. Its ROUGE-1 and ROUGE-2 F1 scores exceed 0.5, demonstrating strong lexical alignment, while the high ROUGE-L value indicates superior sentence-level coherence. BLEU and METEOR scores reinforce this trend, highlighting consistent precision and recall advantages. Moreover, \texttt{Gemma-2} attains the highest exact-match rate, evidencing precise adherence to reference phrasing when contextual alignment permits.
    
            \texttt{GPT-4} follows closely behind, exhibiting balanced and reliable performance across all metrics. Although slight declines are observed at medium and high difficulty levels, it maintains high overall accuracy, achieving BERTScores above 0.9 and EM values comparable to \texttt{Gemma-2}. These results indicate strong factual grounding and effective adherence to the given prompts.
            
            \texttt{Llama-2}, \texttt{Llama-3}, and \texttt{Mistral} form a consistent mid-tier group. Their ROUGE F1 scores are generally lower, primarily due to reduced precision, though their recall remains comparable to that of the leading models. Among these, \texttt{Mistral} exhibits the most stable performance across all difficulty levels, whereas both \texttt{Llama} models experience sharper performance degradation as question complexity increases.
    
            Overall, a consistent performance hierarchy is observed across all evaluation metrics: \texttt{Gemma-2} and \texttt{GPT-4} lead as top performers; \texttt{Llama-2}, \texttt{Llama-3}, and \texttt{Mistral} form a stable middle tier; and \texttt{Gemma} remains the weakest model, often producing responses that lack coherence or contextual grounding. This ranking pattern holds across ROUGE, BLEU, BERTScore and METEOR, indicating robustness in the comparative evaluation.
    
            Among the individual metrics, METEOR yields consistently higher scores than BLEU for all models, as it accounts for stemming, synonymy, and paraphrase matching. This property allows to reward semantically equivalent expressions that BLEU tends to penalize, thereby providing a more lenient yet informative measure of linguistic adequacy. A similar trend is observed for BERTScore, whose values are uniformly high across models due to its semantic sensitivity.

        \subsection{NMISS Performance Evaluation}\label{ch2:sec:results_nmiss}

            Table~\ref{tab:nmissoutperformance} reports the percentage of cases in which NMISS outperforms its traditional counterparts, broken down by model and question difficulty. The results highlight NMISS’s added value across evaluation metrics and task complexities.

            Across all metrics, \texttt{Gemma-2} benefits the most from NMISS, achieving the highest outperformance rates despite having one of the lowest hallucination frequencies. This suggests that its outputs, while factually grounded and semantically coherent, often diverge lexically from the reference answers. Conversely, \texttt{GPT-4} shows smaller NMISS gains, consistent with its strong EM results reported in Section~\ref{ch2:sec:results_rag}, indicating close alignment with the reference in both form and structure.

\begin{figure}[h]
                    \centering
                    \includegraphics[scale = .6]{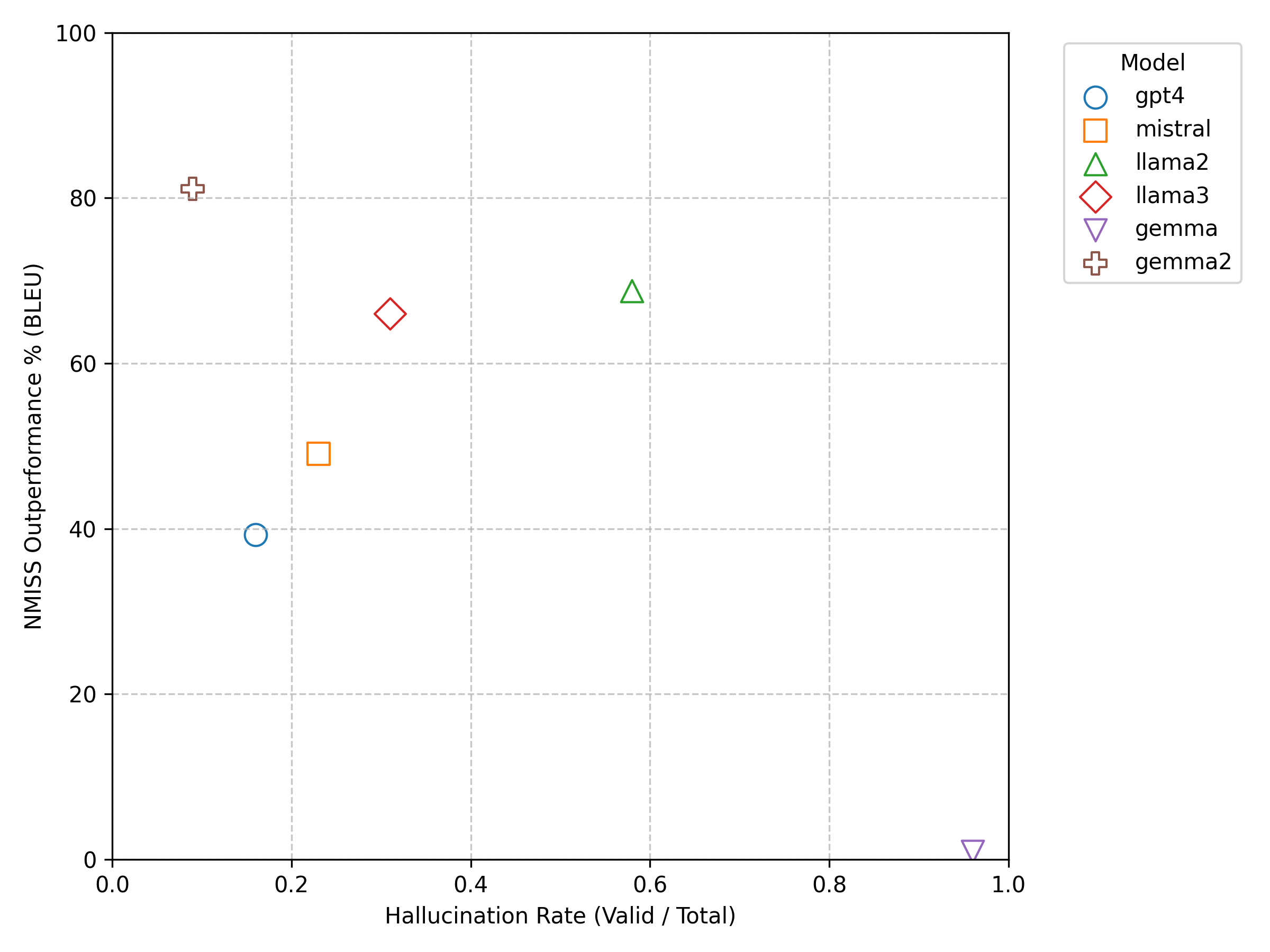}
                    \caption{NMISS Utility vs. Hallucination Rate (BLEU)}
                   \label{fig:nmiss_utility_bleu}
            \end{figure}

\begin{figure}[!h]
                \centering
                \includegraphics[scale = .62]{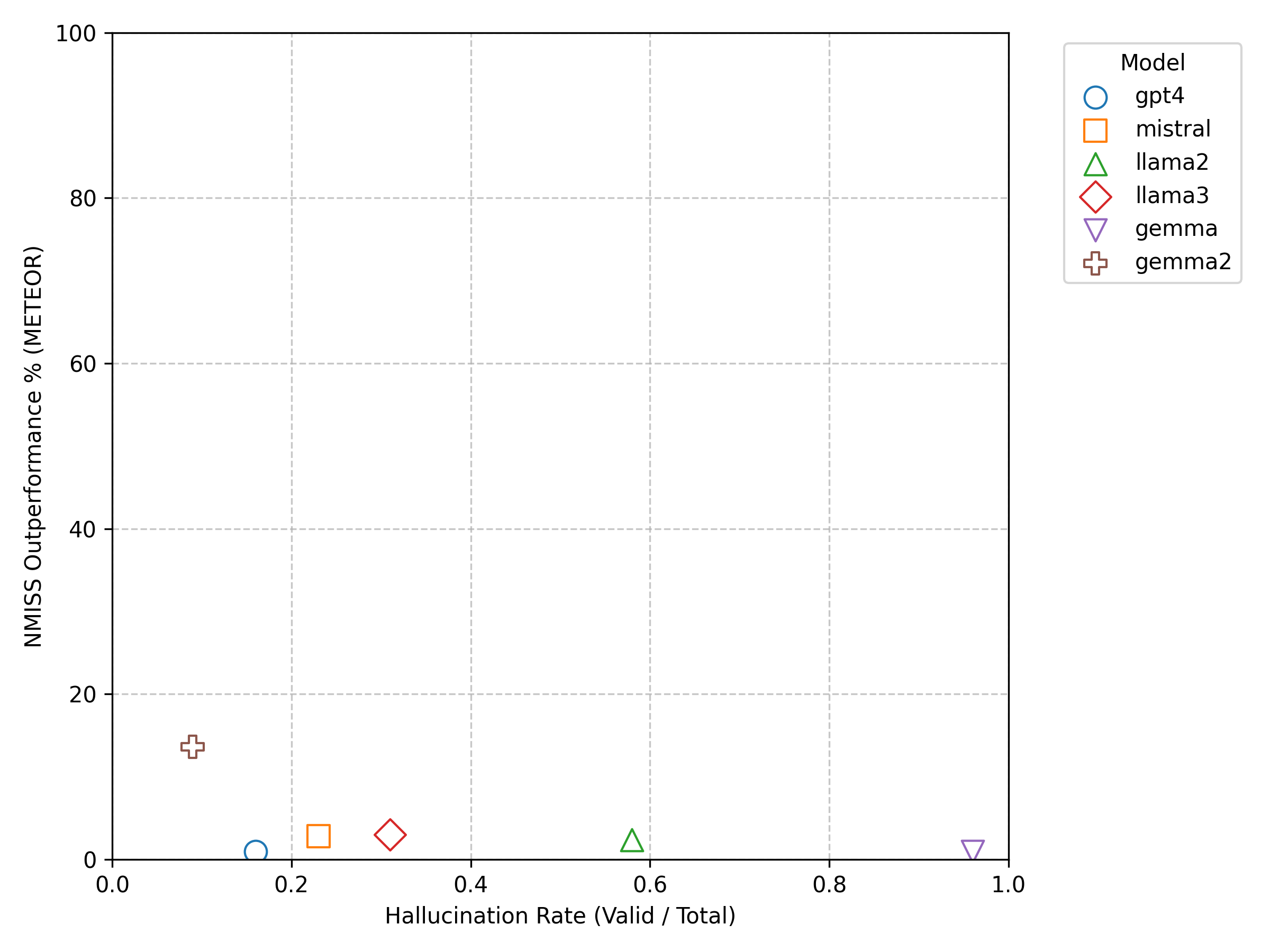}
                \caption{NMISS Utility vs. Hallucination Rate (METEOR)}
               \label{fig:nmiss_utility_meteor}
            \end{figure}

\begin{figure}[!h]
                \centering
                \includegraphics[scale = 0.55]{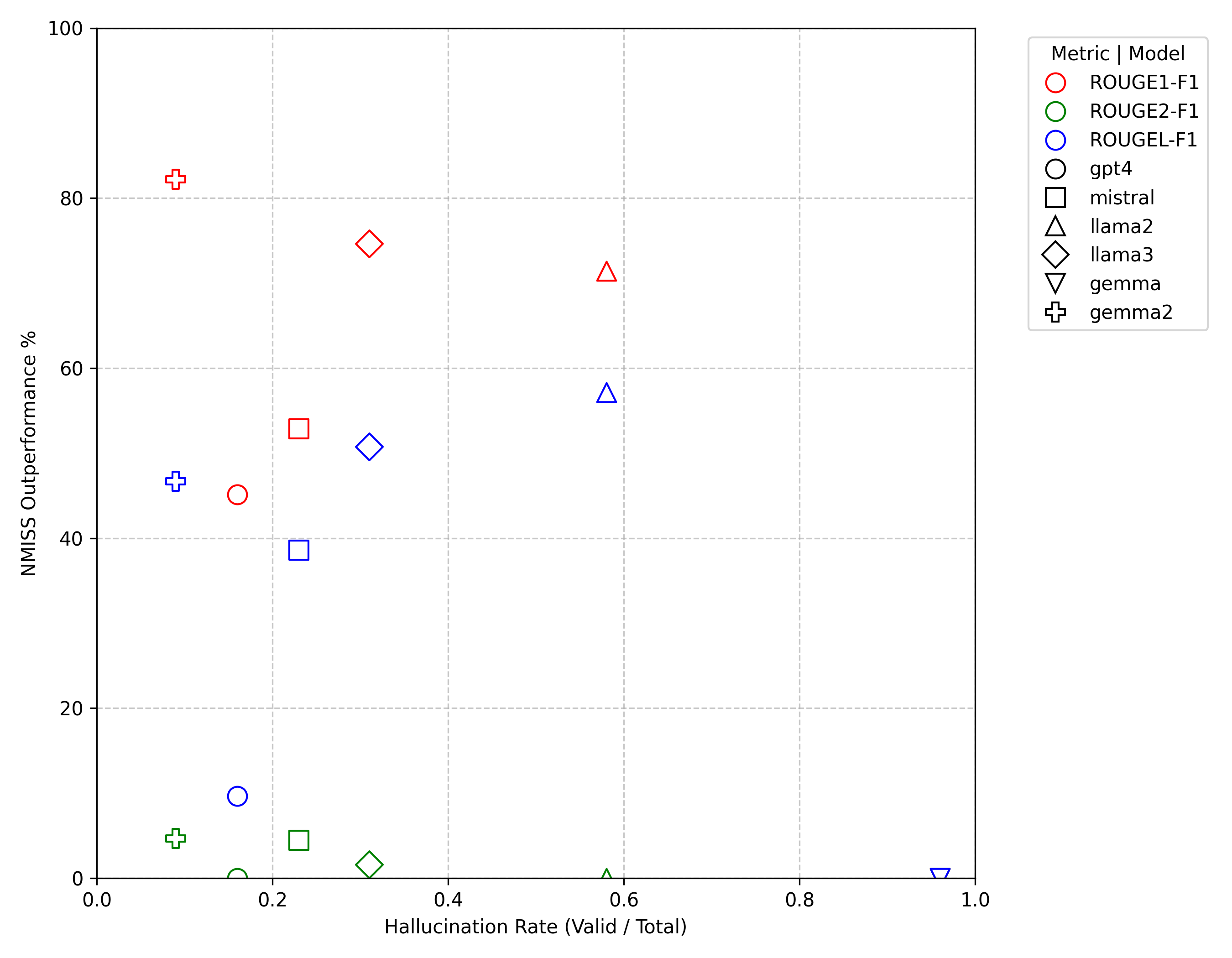}
                \caption{NMISS Utility vs. Hallucination Rate (ROUGE Variants)}
                 \label{fig:nmiss_utility_rouge}
            \end{figure}

            Figures~\ref{fig:nmiss_utility_bleu}--\ref{fig:nmiss_utility_rouge} further examine the relationship between hallucination rate and NMISS outperformance across BLEU, METEOR, and ROUGE variants. For BLEU (Figure~\ref{fig:nmiss_utility_bleu}), a clear positive correlation emerges: models with higher hallucination rates, such as \texttt{LLaMA-2}, \texttt{LLaMA-3}, and \texttt{Mistral}, also exhibit greater NMISS improvements. This pattern supports the view that BLEU penalizes elaborative or contextually grounded completions that deviate from reference phrasing, whereas NMISS compensates by rewarding semantic fidelity. \texttt{Gemma-2} remains an exception, showing substantial NMISS gains despite minimal hallucinations, indicating high-quality, context-enriched responses that standard BLEU scoring undervalues.

                \begin{table}[htbp]
                  \centering
                  \caption{NMISS Outperformance across Models and Question Levels}
                  \begin{threeparttable}
                  \small
                    \begin{tabular}{cccccc}
                    \textbf{Metric} & \textbf{Model} & \textbf{Low} & \textbf{Mid} & \textbf{High} & \textbf{Overall} \\
                    \midrule
                    BLEU  & gemma & 0.00 & 0.00 & 0.00 & 0.00 \\
                    BLEU  & gemma2 & \textbf{87.50} & \textbf{83.33} & \textbf{76.47} & \textbf{82.43} \\
                    BLEU  & gpt4  & 50.00 & 28.57 & 40.00 & 39.52 \\
                    BLEU  & llama2 & 66.67 & 75.00 & 63.64 & 68.43 \\
                    BLEU  & llama3 & 66.67 & 60.00 & 73.33 & 66.67 \\
                    BLEU  & mistral & 50.00 & 45.83 & 52.63 & 49.49 \\
                    \midrule
                    METEOR & gemma & 0.00  & 0.00  & 0.00  & 0.00 \\
                    METEOR & gemma2 & 0.00  & \textbf{21.43} & \textbf{15.00} & \textbf{12.14} \\
                    METEOR & gpt4  & 0.00  & 0.00  & 0.00  & 0.00 \\
                    METEOR & llama2 & 0.00  & 6.25  & 0.00  & 2.08 \\
                    METEOR & llama3 & \textbf{4.76} & 0.00  & 4.35  & 3.04 \\
                    METEOR & mistral & 0.00  & 3.70  & 3.70  & 2.47 \\
                    \midrule
                    ROUGE-1 F1 & gemma & 0.00  & 0.00  & 0.00  & 0.00 \\
                    ROUGE-1 F1 & gemma2 & \textbf{81.82} & \textbf{85.71} & \textbf{80.00} & \textbf{82.51} \\
                    ROUGE-1 F1 & gpt4  & 71.43 & 28.57 & 41.18 & 47.06 \\
                    ROUGE-1 F1 & llama2 & 75.00 & 75.00 & 64.29 & 71.43 \\
                    ROUGE-1 F1 & llama3 & 71.43 & 78.26 & 73.91 & 74.53 \\
                    ROUGE-1 F1 & mistral & 41.18 & 57.69 & 55.56 & 51.47 \\
                    \midrule
                    ROUGE-2 F1 & gemma & 0.00  & 0.00  & 0.00  & 0.00 \\
                    ROUGE-2 F1 & gemma2 & \textbf{9.09} & \textbf{7.69} & 0.00  & \textbf{5.59} \\
                    ROUGE-2 F1 & gpt4  & 0.00  & 0.00  & 0.00  & 0.00 \\
                    ROUGE-2 F1 & llama2 & 0.00  & 0.00  & 0.00  & 0.00 \\
                    ROUGE-2 F1 & llama3 & 0.00  & 0.00  & \textbf{4.76} & 1.59 \\
                    ROUGE-2 F1 & mistral & 6.67  & 7.69  & 0.00  & 4.79 \\
                    \midrule
                    ROUGE-L F1 & gemma & 0.00  & 0.00  & 0.00  & 0.00 \\
                    ROUGE-L F1 & gemma2 & 36.36 & 50.00 & 50.00 & 45.45 \\
                    ROUGE-L F1 & gpt4  & 0.00  & 14.29 & 11.76 & 8.68 \\
                    ROUGE-L F1 & llama2 & \textbf{66.67} & \textbf{56.25} & 50.00 & \textbf{57.64} \\
                    ROUGE-L F1 & llama3 & 52.38 & 43.48 & \textbf{56.52} & 50.79 \\
                    ROUGE-L F1 & mistral & 29.41 & 50.00 & 33.33 & 37.58 \\
                    \bottomrule
                    \bottomrule
                    \end{tabular}%
                     \begin{tablenotes}
                        \footnotesize
                        \item \textbf{Note.} Scores represent the percentage of hallucination cases where NMISS outperformed the corresponding traditional metrics, expressed as the ratio of improved cases to valid hallucinations. Bold values highlight the highest scores per metric. “Overall” is the average across all questions difficulty. NMISS outperformance is computed over valid hallucination cases (excluding degenerate scores of 0 or 0.99).
                        \end{tablenotes}
                      \end{threeparttable}
                 \label{tab:nmissoutperformance}%
                \end{table}%

            In the case of METEOR (Figure~\ref{fig:nmiss_utility_meteor}), the effect is less pronounced. This metric already incorporates stemming, synonymy, and paraphrase matching and naturally captures a larger share of semantic variability between system and reference outputs. Consequently, the set of non-matching tokens is comparatively small, reducing the influence of the contextual component in NMISS. Under these conditions, NMISS primarily refines borderline cases rather than producing large-scale adjustments. Nonetheless, \texttt{Gemma-2} and \texttt{LLaMA-3} still exhibit meaningful gains, suggesting that their elaborations are contextually grounded even when lexical overlap is already high.

            ROUGE analysis (Figure~\ref{fig:nmiss_utility_rouge}) reveals more varied patterns across its variants. ROUGE-1 largely mirrors BLEU’s behavior, with higher hallucination rates correlating with stronger NMISS improvements. ROUGE-L shows a weaker but still positive relationship, while ROUGE-2 remains mostly insensitive to NMISS adjustments, reflecting the rigidity of bigram-based evaluation in capturing nuanced or rephrased content.

            Overall, NMISS enhances evaluation in two complementary scenarios: (i) for models that hallucinate more frequently, by distinguishing spurious additions from contextually valid elaborations; and (ii) for high-performing models such as \texttt{Gemma-2}, by recognizing semantically grounded deviations that traditional metrics undervalue. These findings confirm NMISS’s complementary role and robustness across varying task difficulty levels and model qualities. Detailed average NMISS and baseline scores for BLEU, METEOR, and ROUGE are reported in Tables~\ref{tab:bleu_nmiss_results}--\ref{tab:rouge_nmiss_results_L} in the Appendix.

   \section{Conclusions}\label{ch2:sec:conclusions}

       This study investigates hallucination behavior in LLMs from a dual perspective: mitigation through a RAG architecture and evaluation refinement via the proposed NMISS. The experimental framework focuses on a question-answering task over Italian health news articles, simulating user–agent interactions in a sensitive knowledge domain. Six models—\texttt{Gemma}, \texttt{Gemma-2}, \texttt{GPT-4}, \texttt{LLaMA-2}, \texttt{LLaMA-3}, and \texttt{Mistral}—are evaluated in a zero-shot configuration with a fixed prompt template to ensure consistent generation behavior across systems.

       The analysis shows that \texttt{Gemma-2} and \texttt{GPT-4} achieve the strongest overall performance under traditional metrics such as BLEU, ROUGE, and METEOR. \texttt{Gemma-2} demonstrates balanced precision and recall across all difficulty levels, making it a promising candidate for healthcare-related applications. \texttt{GPT-4}, while slightly behind on surface-based metrics, attains the highest EM accuracy, reflecting its strict prompt adherence and literal generation style. Mid-tier models—\texttt{LLaMA-2}, \texttt{LLaMA-3}, and \texttt{Mistral}—achieve comparable recall but lower precision, producing contextually relevant yet lexically diverse completions that traditional metrics tend to undervalue.

       To address this evaluative limitation, the NMISS framework is introduced to reward contextual alignment by incorporating information retrieved from the original input document. Focusing on non-hallucinated responses, the analysis quantifies the proportion of cases in which NMISS outperforms standard metrics, yielding several key findings. First, NMISS effectively identifies hidden semantic value in outputs overlooked by conventional metrics. \texttt{Gemma-2}, despite already leading in raw performance, achieves the highest NMISS outperformance across BLEU, ROUGE-1, and METEOR, demonstrating that NMISS can recognize semantically rich yet lexically divergent answers that remain contextually faithful. Second, although \texttt{GPT-4} performs strongly overall, its NMISS improvement is modest. This reflects its rigid zero-shot generation behavior, which closely mirrors the prompt and minimizes lexical variation, leaving less room for NMISS to detect additional contextual contributions.

        Conversely, mid-tier models exhibit substantial NMISS gains, particularly for BLEU and ROUGE-1. Their responses, while more syntactically variable, frequently contain relevant contextual cues that overlap-based metrics penalize but NMISS rewards. This suggests that NMISS captures a compensatory strategy: when these models struggle to respond directly, they often enrich their outputs with contextually grounded elaborations that enhance informativeness.

        A positive correlation emerges between hallucination rate and NMISS outperformance for BLEU and ROUGE-1, reinforcing the idea that NMISS excels in borderline cases where literal matching fails to capture valid, context-driven reasoning. Notably, \texttt{Gemma-2} remains an exception, showing that even factually accurate responses can benefit from a more context-sensitive evaluation.

        Overall, the findings demonstrate that while RAG architectures effectively mitigate hallucinations by grounding LLM outputs, NMISS complements this process by refining how output quality is assessed. It compensates for the underestimation introduced by traditional overlap-based metrics, which often assign lower scores to responses that are semantically correct but lexically divergent. By rewarding contextually aligned and informative outputs, NMISS provides a fairer evaluation of model performance. This dual contribution is particularly relevant for high-stakes domains such as healthcare, where contextual accuracy and informativeness often outweigh surface-level similarity.

        In conclusion, NMISS represents a valuable addition to the LLM evaluation toolkit. It enhances fairness and interpretability in performance assessment, particularly for models producing rich and context-aware responses. While top-performing models benefit from NMISS refinement, mid-tier models gain even more, as the metric recovers value otherwise overlooked by standard approaches. 
        
        Future work could extend NMISS to multilingual and multi-hop question answering, adaptive fine-tuning pipelines, and reinforcement learning frameworks that align model outputs with context-based quality objectives. Beyond the Italian healthcare domain, the combined RAG–NMISS methodology generalizes to other languages and knowledge domains where factual grounding and contextual relevance are essential.

    \clearpage
    \section{Appendix}
    
                \begin{table}[H]
                 \centering
                 \begin{threeparttable}
                     \caption{BLEU versus BLEU NMISS scores}
                     \small
                    \begin{tabular}{c|ccc|ccc}
                        \multicolumn{1}{c}{} & \multicolumn{3}{c}{\textbf{BLEU score}} & \multicolumn{3}{c}{\textbf{BLEU NMISS score}} \\
                        \midrule
                        \midrule
                        \textbf{Model} & \textbf{LOW} & \textbf{MID} & \textbf{HIGH} & \textbf{LOW} & \textbf{MID} & \textbf{HIGH} \\
                        \midrule
                        gpt4  & 0.650 & 0.468 & 0.473 & \textbf{0.653} & \textbf{0.469} & \textbf{0.484} \\
                        llama2 & 0.122 & 0.072 & 0.067 & \textbf{0.151} & \textbf{0.098} & \textbf{0.090} \\
                        llama3 & 0.152 & 0.208 & 0.125 & \textbf{0.170} & \textbf{0.259} & \textbf{0.173} \\
                        mistral & 0.272 & 0.263 & 0.191 & \textbf{0.291} & \textbf{0.291} & \textbf{0.225} \\
                        gemma & \textbf{0.001} & 0.000 & 0.047 & 0.001 & \textbf{0.011} & \textbf{0.049} \\
                        gemma2 & 0.654 & 0.578 & 0.604 & 0.668 & \textbf{0.626} & \textbf{0.664} \\
                        \bottomrule
                    \end{tabular}
                     \begin{tablenotes}
                        \item {Note: Average scores. Higher values are highlighted in bold.}
                        \end{tablenotes}
                        \end{threeparttable}
                      \label{tab:bleu_nmiss_results}
                \end{table}%

                \begin{table}[h]
                    \centering
                    \begin{threeparttable}
                     \caption{METEOR versus METEOR NMISS scores}
                     \small
                    \begin{tabular}{c|ccc|ccc}
                    \multicolumn{1}{c}{} & \multicolumn{3}{c}{\textbf{METEOR score}} & \multicolumn{3}{c}{\textbf{METEOR NMISS score}} \\
                    \midrule
                    \midrule
                    \textbf{Model} & \textbf{LOW} & \textbf{MID} & \textbf{HIGH} & \textbf{LOW} & \textbf{MID} & \textbf{HIGH} \\
                    \midrule
                    gpt4  & \multicolumn{1}{r}{\textbf{0.714}} & \textbf{0.526} & 0.565 & 0.714 & 0.526 & \textbf{0.567} \\
                    llama2 & 0.443 & 0.325 & 0.291 & \textbf{0.453} & \textbf{0.330} & \textbf{0.310} \\
                    llama3 & 0.415 & 0.442 & 0.336 & \textbf{0.429} & \textbf{0.455} & \textbf{0.355} \\
                    mistral & 0.431 & 0.485 & 0.451 & \textbf{0.436} & \textbf{0.488} & \textbf{0.456} \\
                    gemma & 0.010 & 0.012 & 0.073 & \textbf{0.011} & \textbf{0.019} & \textbf{0.080} \\
                    gemma2 & 0.718 & 0.675 & 0.724 & \textbf{0.723} & \textbf{0.682} & \textbf{0.731} \\
                    \bottomrule
                    \end{tabular}%
                      \begin{tablenotes}
                        \item {Note: Average scores. Higher values are highlighted in bold.}
                        \end{tablenotes}
                    \end{threeparttable}
                  \label{tab:meteor_nmiss_results}
                \end{table}%
    
                \begin{table}[h]
                  \centering
                  \begin{threeparttable}
                   \caption{ROUGE-1 versus ROUGE-1 NMISS scores}
                   \small
                    \begin{tabular}{l|l|ccc|ccc}
                    \multicolumn{8}{c}{\textbf{ROUGE-1 versus ROUGE-1 NMISS}} \\
                    \midrule
                    \midrule
                    \multicolumn{1}{l}{Level} & Model & \multicolumn{1}{l}{P} & \multicolumn{1}{l}{R} & \multicolumn{1}{l}{F1} & \multicolumn{1}{l}{NMISS\_P} & \multicolumn{1}{l}{NMISS\_R} & \multicolumn{1}{l}{NMISS\_F1} \\
                    \midrule
                    LOW   & gpt4  & 0.721 & \textbf{0.740} & 0.718 & \textbf{0.742} & 0.740 & \textbf{0.728} \\
                    LOW   & llama2 & 0.201 & 0.712 & 0.282 & \textbf{0.342} & \textbf{0.717} & \textbf{0.421} \\
                    LOW   & llama3 & 0.312 & 0.557 & 0.338 & \textbf{0.462} & \textbf{0.564} & \textbf{0.433} \\
                    LOW   & mistral & 0.434 & 0.489 & 0.421 & \textbf{0.531} & \textbf{0.492} & \textbf{0.466} \\
                    LOW   & gemma & 0.028 & 0.018 & 0.014 & \textbf{0.043} & 0.018 & \textbf{0.021} \\
                    LOW   & gemma2 & 0.793 & 0.727 & 0.738 & \textbf{0.858} & \textbf{0.731} & \textbf{0.758} \\
                    \midrule
                    MID   & gpt4  & 0.552 & 0.529 & 0.530 & \textbf{0.603} & \textbf{0.531} & \textbf{0.539} \\
                    MID   & llama2 & 0.153 & 0.510 & 0.211 & \textbf{0.280} & \textbf{0.511} & \textbf{0.308} \\
                    MID   & llama3 & 0.356 & 0.605 & 0.384 & \textbf{0.558} & \textbf{0.610} & \textbf{0.488} \\
                    MID   & mistral & 0.434 & 0.547 & 0.447 & \textbf{0.547} & \textbf{0.550} & \textbf{0.498} \\
                    MID   & gemma & 0.017 & 0.018 & 0.015 & \textbf{0.049} & \textbf{0.022} & \textbf{0.026} \\
                    MID   & gemma2 & 0.727 & 0.698 & 0.688 & \textbf{0.877} & \textbf{0.701} & \textbf{0.735} \\
                    \midrule
                    HIGH  & gpt4  & 0.585 & 0.603 & 0.582 & \textbf{0.647} & \textbf{0.604} & \textbf{0.611} \\
                    HIGH  & llama2 & 0.156 & 0.477 & 0.211 & \textbf{0.301} & \textbf{0.485} & \textbf{0.330} \\
                    HIGH  & llama3 & 0.271 & 0.432 & 0.295 & \textbf{0.491} & \textbf{0.452} & \textbf{0.401} \\
                    HIGH  & mistral & 0.364 & 0.537 & 0.396 & \textbf{0.499} & \textbf{0.539} & \textbf{0.455} \\
                    HIGH  & gemma & 0.084 & 0.084 & 0.076 & \textbf{0.111} & \textbf{0.091} & \textbf{0.095} \\
                    HIGH  & gemma2 & 0.706 & 0.755 & 0.715 & \textbf{0.880} & \textbf{0.758} & \textbf{0.790} \\
                    \bottomrule
                    \end{tabular}%
                     \begin{tablenotes}
                        \item {Note: Average scores. Higher values are highlighted in bold.}
                        \end{tablenotes}
                        \end{threeparttable}
                  \label{tab:rouge_nmiss_results_1}%
                \end{table}%
    
                \begin{table}[h]
                  \centering
                  \begin{threeparttable}
                  \caption{ROUGE-2 versus ROUGE-2 NMISS scores}
                  \small
                    \begin{tabular}{c|c|ccc|ccc}
                    \multicolumn{8}{c}{\textbf{ROUGE-2 versus ROUGE-2 NMISS}} \\
                    \midrule
                    \midrule
                    \multicolumn{1}{l}{Level} & \multicolumn{1}{l|}{Model} & \multicolumn{1}{l}{P} & \multicolumn{1}{l}{R} & \multicolumn{1}{l}{F1} & \multicolumn{1}{l}{NMISS\_P} & \multicolumn{1}{l}{NMISS\_R} & \multicolumn{1}{l}{NMISS\_F1} \\
                    \midrule
                    LOW   & gpt4  & \textbf{0.665} & \textbf{0.682} & \textbf{0.669} & 0.665 & 0.682 & 0.669 \\
                    LOW   & llama2 & \textbf{0.131} & \textbf{0.568} & \textbf{0.191} & 0.131 & 0.568 & 0.191 \\
                    LOW   & llama3 & 0.218 & 0.449 & 0.243 & \textbf{0.223} & \textbf{0.450} & \textbf{0.245} \\
                    LOW   & mistral & 0.337 & 0.409 & 0.340 & \textbf{0.338} & \textbf{0.409} & \textbf{0.340} \\
                    LOW   & gemma & \textbf{0.020} & \textbf{0.005} & \textbf{0.008} & 0.020 & 0.005 & 0.008 \\
                    LOW   & gemma2 & 0.730 & 0.687 & 0.692 & \textbf{0.735} & \textbf{0.688} & \textbf{0.693} \\
                    \midrule
                    MID   & gpt4  & \textbf{0.490} & \textbf{0.486} & \textbf{0.485} & 0.490 & 0.486 & 0.485 \\
                    MID   & llama2 & 0.084 & 0.353 & 0.120 & \textbf{0.085} & \textbf{0.353} & \textbf{0.121} \\
                    MID   & llama3 & 0.270 & 0.472 & 0.284 & \textbf{0.273} & \textbf{0.473} & \textbf{0.286} \\
                    MID   & mistral & 0.326 & 0.453 & 0.349 & \textbf{0.333} & \textbf{0.454} & \textbf{0.351} \\
                    MID   & gemma & \textbf{0.004} & \textbf{0.003} & \textbf{0.004} & 0.004 & 0.003 & 0.004 \\
                    MID   & gemma2 & 0.662 & 0.637 & 0.632 & \textbf{0.665} & \textbf{0.638} & \textbf{0.633} \\
                    \midrule
                    HIGH  & gpt4  & 0.493 & 0.505 & 0.496 & \textbf{0.494} & \textbf{0.505} & \textbf{0.496} \\
                    HIGH  & llama2 & 0.075 & 0.280 & 0.106 & \textbf{0.076} & \textbf{0.281} & \textbf{0.107} \\
                    HIGH  & llama3 & 0.155 & 0.295 & 0.179 & \textbf{0.163} & \textbf{0.296} & \textbf{0.181} \\
                    HIGH  & mistral & 0.242 & 0.412 & 0.281 & \textbf{0.244} & \textbf{0.413} & \textbf{0.282} \\
                    HIGH  & gemma & \textbf{0.067} & \textbf{0.056} & \textbf{0.060} & 0.067 & 0.056 & 0.060 \\
                    HIGH  & gemma2 & 0.635 & 0.664 & 0.641 & \textbf{0.639} & \textbf{0.664} & \textbf{0.642} \\
                    \bottomrule
                    \end{tabular}%
                      \begin{tablenotes}
                        \item {Note: Average scores. Higher values are highlighted in bold.}
                        \end{tablenotes}
                    \end{threeparttable}
                  \label{tab:rouge_nmiss_results_2}%
                \end{table}%

                \begin{table}[h]
                  \centering
                      \begin{threeparttable}
                    \caption{ROUGE-L versus ROUGE-L NMISS scores}
                    \small
                    \begin{tabular}{c|c|ccc|ccc}
                    \multicolumn{8}{c}{\textbf{ROUGE-L versus ROUGE-L NMISS}} \\
                    \midrule
                    \midrule
                    \multicolumn{1}{c}{Level} & Model & P     & R     & \multicolumn{1}{c}{F1} & NMISS\_P & NMISS\_R & NMISS\_F1 \\
                    \midrule
                    LOW   & gpt4  & 0.709 & \textbf{0.722} & 0.705 & \textbf{0.719} & 0.722 & \textbf{0.707} \\
                    LOW   & llama2 & 0.171 & 0.657 & 0.245 & \textbf{0.206} & \textbf{0.658} & \textbf{0.284} \\
                    LOW   & llama3 & 0.274 & 0.513 & 0.301 & \textbf{0.337} & \textbf{0.514} & \textbf{0.338} \\
                    LOW   & mistral & 0.390 & 0.454 & 0.385 & \textbf{0.440} & \textbf{0.456} & \textbf{0.406} \\
                    LOW   & gemma & 0.023 & \textbf{0.016} & 0.012 & \textbf{0.033} & 0.016 & \textbf{0.018} \\
                    LOW   & gemma2 & 0.779 & 0.719 & 0.729 & \textbf{0.805} & \textbf{0.721} & \textbf{0.735} \\
                    \midrule
                    MID   & gpt4  & 0.538 & 0.519 & 0.519 & \textbf{0.575} & \textbf{0.520} & \textbf{0.523} \\
                    MID   & llama2 & 0.129 & 0.460 & 0.180 & \textbf{0.165} & \textbf{0.461} & \textbf{0.214} \\
                    MID   & llama3 & 0.321 & \textbf{0.555} & 0.345 & \textbf{0.373} & \textbf{0.555} & \textbf{0.377} \\
                    MID   & mistral & 0.377 & 0.500 & 0.398 & \textbf{0.413} & 0.502 & \textbf{0.413} \\
                    MID   & gemma & 0.011 & 0.013 & 0.009 & \textbf{0.022} & \textbf{0.014} & \textbf{0.014} \\
                    MID   & gemma2 & 0.702 & 0.673 & 0.665 & \textbf{0.743} & \textbf{0.673} & \textbf{0.677} \\
                    \midrule
                    HIGH  & gpt4  & 0.551 & 0.562 & 0.546 & \textbf{0.573} & \textbf{0.562} & \textbf{0.554} \\
                    HIGH  & llama2 & 0.120 & 0.406 & 0.167 & \textbf{0.159} & \textbf{0.408} & \textbf{0.201} \\
                    HIGH  & llama3 & 0.227 & 0.382 & 0.252 & \textbf{0.293} & \textbf{0.389} & \textbf{0.285} \\
                    HIGH  & mistral & 0.316 & \textbf{0.489} & 0.353 & \textbf{0.356} & 0.489 & \textbf{0.372} \\
                    HIGH  & gemma & 0.079 & 0.079 & 0.072 & \textbf{0.092} & \textbf{0.084} & \textbf{0.082} \\
                    HIGH  & gemma2 & 0.677 & \textbf{0.718} & 0.684 & \textbf{0.707} & 0.718 & \textbf{0.698} \\
                    \bottomrule
                    \end{tabular}%
                      \begin{tablenotes}
                        \item {Note: Average scores. Higher values are highlighted in bold.}
                        \end{tablenotes}
                    \end{threeparttable}
                  \label{tab:rouge_nmiss_results_L}%
                \end{table}%

\clearpage
\bibliographystyle{plainnat} 
\bibliography{biblio}

\end{document}